\documentclass[runningheads]{llncs}

\usepackage{eccv}

\usepackage{eccvabbrv}

\usepackage{graphicx}
\usepackage{booktabs}
\usepackage{multirow}
\usepackage{makecell}
\usepackage{pifont}
\usepackage{colortbl}
\usepackage{xcolor}
\allowdisplaybreaks[4]
\usepackage[linesnumbered,ruled,vlined]{algorithm2e}

\usepackage{lipsum}
\usepackage{etoolbox}

\definecolor{tablecolor}{HTML}{CEE9DC}
\definecolor{cifarcolor}{HTML}{F3E6CA}
\definecolor{tinycolor}{HTML}{BCC6DD}
\definecolor{domaincolor}{HTML}{EFD6D1}

\newtheorem{assumption}{Assumption}

\usepackage[breaklinks,colorlinks,citecolor=eccvblue]{hyperref}

\hypersetup{linktoc=all}

\usepackage{orcidlink}

\makeatletter
\let\maketitle@orig\maketitle
\renewcommand{\maketitle}{%
  \begingroup
    \let\addcontentsline\@gobblethree
    \let\addtocontents\@gobbletwo
    \maketitle@orig
  \endgroup
}
\makeatother

\begin{document}

\setcounter{tocdepth}{-1}
\addtocontents{toc}{\protect\setcounter{tocdepth}{-1}}

\title{Fisher-Routed Mixture of Experts for Federated Class-Incremental Learning} 

\titlerunning{\textsc{FedFMX}}

\author{Wenhao Yuan\inst{1}\orcidlink{0009-0001-6625-7496} \and
Chenchen Lin\inst{2}\orcidlink{0009-0002-8473-6068} \and
Jian Chen\inst{1}\orcidlink{0000-0002-4570-2271} \and
Jinfeng Xu\inst{1}\orcidlink{0009-0001-7876-3740} \and
Zewei Liu\inst{1}\orcidlink{0000-0003-1038-3204} \and
Edith Cheuk Han Ngai\inst{1,\thanks{Corresponding author.}}\orcidlink{0000-0002-3454-8731} 
}

\authorrunning{W.~Yuan et al.}

\institute{Department of Electrical and Computer Engineering, The University of Hong Kong, Hong Kong SAR, China \and
School of Artificial Intelligence, Sun Yat-sen University, China. \\
\email{wenhao.yuan@connect.hku.hk}, \email{chngai@eee.hku.hk}\\
}

\maketitle

\begin{abstract}
Federated Learning (FL) emerged as a promising distributed machine learning paradigm. However, extending FL to the class incremental learning scenarios introduces unique challenges: 1) Capacity conflict and catastrophic forgetting from the shared model overloading, 2) Heterogeneity from Non-Independent and Identically Distributed (Non-IID) data, and 3) Synchronized class misalignment. In this paper, we propose \textbf{F}isher-Routed \textbf{M}i\textbf{X}ture of Experts for \textbf{Fed}erated Class-Incremental Learning (\textsc{FedFMX}), a novel framework to address these challenges via adaptive expert specialization across clients. The crucial insight is to route each sample to an expert subset that jointly optimizes knowledge acquisition and retention. Specifically, we introduce a Fisher-Routed Expert Scoring (FRES) module to estimate expert importance via Fisher-based stability cost and gradient-based plasticity gain. Then, we design an Adaptive Expert Selection (AES) module by quantifying marginal contributions for adaptive expert subset determination. Finally, by the routing-aware regularization (RAR), we achieve load balance and efficient FL training. We theoretically prove the $\mathcal{O}(T^{-1})$ convergence rate. Extensive experiments on multiple benchmarks compared with state-of-the-art methods demonstrate the superiority of \textsc{FedFMX}.
\end{abstract}

\section{Introduction} \label{sec:intro}

\textit{Federated Learning} (FL)~\cite{mcmahan2017communication, kairouz2021advances, li2020federated, fan2025ten} has emerged as a promising distributed learning paradigm, enabling multiple geographically decentralized clients to jointly learn a shared global model without exposing their raw data. However, conventional FL approaches typically assume that \textit{the label distribution remains static and consistent during the training process}~\cite{zhang2025pfedmxf, liang2025class}, which rarely holds in real-world applications~\cite{feng2025cgofed, geng2020recent, yang2024federated}. In practice, due to dynamic contexts, client data distributions evolve, and clients continuously receive new classes of data with previously unseen classes~\cite{dong2022federated, shi2021efficient}, thus invalidating the aforementioned ideal assumption. Further, given that clients are generally resource-constrained with limited storage and computational capacity~\cite{li2025resource}, maintaining a complete history of past data is infeasible, which exacerbates the risk of catastrophic forgetting~\cite{babakniya2023data, yu2024overcoming}.

To overcome the limitation of the static label distribution assumption, recent efforts have advanced \textit{Federated Class-Incremental Learning} (FCIL) paradigm~\cite{feng2025cgofed, dong2023no}, enabling clients to incrementally learn novel classes while collaboratively updating the global model~\cite{liang2025class}. However, the intersection of the federated and incremental paradigms introduces several intrinsic and crucial challenges~\cite{dong2022federated, zhang2025pfedmxf}: (\lowercase\expandafter{\romannumeral1}) \textit{Capacity Conflict and Catastrophic Forgetting}: Since the shared global model is incrementally overloaded with heterogeneous data, the learned representations are overwritten, leading to catastrophic forgetting and degraded generalization~\cite{you2025adaptive, ke2025task}; (\lowercase\expandafter{\romannumeral2}) \textit{Statistical Heterogeneity}: Clients possess dynamically evolving Non-IID data, resulting in the gradient direction inconsistency and biased local updates, impairing global convergence~\cite{guo2024pilora, gao2024fedprok}; (\lowercase\expandafter{\romannumeral3}) \textit{Synchronized Class Misalignment}: Clients encounter newly introduced classes at various phases, resulting in temporally inconsistent label spaces, exacerbating model drifting~\cite{zhang2025pfedmxf, guan2025stsa}. 

To alleviate catastrophic forgetting in FCIL, prior works adopt global regularization, memory replay, or rehearsal-free techniques~\cite{dong2022federated, yu2024overcoming, sun2025rehearsal}. However, these methods default the global model to a monolithic entity, without fine-grained mechanisms to modulate stability and plasticity, particularly pronounced in heterogeneous scenarios that require differentiated treatment across model parameters. Drawing from these analyses, we raise \textbf{Key Question \uppercase\expandafter{\romannumeral1}}: \textit{How to balance stability and plasticity in FCIL such that the global model retains prior knowledge while adapting to incrementally emerging classes across heterogeneous clients?} Another fundamental challenge arises from the compounding effect of statistical heterogeneity and temporal class drifting. Clients dynamically receive novel classes at different phases, leading to the \textit{synchronized class misalignment} problem, where the semantic inconsistency of class distributions across clients impairs convergence stability and generalization~\cite{dong2022federated, yu2024overcoming, zhang2025pfedmxf}. While recent works introduce personalized methods to mitigate heterogeneity~\cite{yi2026pFedMoE}, overlooking tailoring collaboration based on clients' evolving task profiles, limits the scalability. Then, we pose \textbf{Key Question \uppercase\expandafter{\romannumeral2}:} \textit{How to achieve stable and effective collaboration in FCIL under client heterogeneity and temporally divergent class updates, ensuring robust and efficient training?}

To address these issues, we present a novel solution, \textbf{F}isher-Routed \textbf{M}i\textbf{X}ture of Experts for \textbf{Fed}erated Class-Incremental Learning (\textsc{FedFMX}), which introduces the expert modularity and adaptive activation and routing to disentangle evolving and heterogeneous client dynamics. For question \textbf{\uppercase\expandafter{\romannumeral1}}, we design a \textit{Fisher-Routed Expert Scoring} (FRES) module to estimate suitability by quantifying the stability cost and plasticity gain of each expert through Fisher information in \S~\ref{fisher_route}, enabling context-aware routing. For question \textbf{\uppercase\expandafter{\romannumeral2}}, we introduce an \textit{Adaptive Expert Selection} (AES) module in \S~\ref{Adp_Expert_Selection}, which formulates expert subset selection as a cooperative game and activates experts dynamically via the marginal contribution to stability–plasticity trade-offs. To enhance routing robustness and mitigate biased expert utilization, we present a \textit{routing-aware regularization} (RAR) scheme in \S~\ref{regularization}. Our main contributions are summarized as follows:
\begin{itemize}
\item We identify the significant challenges in FCIL and propose \textsc{FedFMX} to mitigate catastrophic forgetting and heterogeneity by routing each sample to a dynamical expert subset, achieving fine-grained specialization and robust knowledge integration.

\item We design the FRES module to quantify the expert suitability via stability-plasticity trade-offs, where we evaluate the marginal contribution of each expert and determine adaptive expert subsets, enabling efficient expert collaboration across temporally misaligned classes. Through the RAR strategy, we promote stable and fair expert utilization.

\item Extensive experiments on CIFAR-10, CIFAR-100~\cite{krizhevsky2009learning}, and Tiny-ImageNet~\cite{le2015tiny} datasets validate the effectiveness of \textsc{FedFMX}, showing consistent improvements over state-of-the-art methods.
\end{itemize}

\section{Related Work}

\subsection{Federated Class-Incremental Learning}
\textit{Federated Class-Incremental Learning} (FCIL)~\cite{dong2022federated} extends the FL paradigm to continual learning with training while incrementally acquiring new classes~\cite{gao2024fedprok, li2024sr}, which poses crucial challenges, such as asynchronous task arrivals~\cite{zhang2025pfedmxf, ke2025task, li2025cross} and Non-IID data~\cite{wang2024digital, liang2025class}, leading to catastrophic forgetting and unbalanced updates~\cite{yu2024overcoming, you2025adaptive, feng2025cgofed}. To solve these issues, regularization-based methods~\cite{feng2025cgofed, tan2024fl} are proposed to constrain updates to preserve previously acquired knowledge. For instance, CGoFed~\cite{feng2025cgofed} and LGA~\cite{dong2023no} include group-aware regularization to alleviate catastrophic forgetting. Meanwhile, many approaches adopt replay-based strategies~\cite{li2025re, ke2025task, sun2025rehearsal, nori2025autoencoderbased} to maintain a local memory buffer or a synthetic generator to replay past samples. In contrast, rehearsal-free methods~\cite{sun2025rehearsal, li2025fedssi, he2025cl} eliminate memory storage by relying on parameter isolation or prompt tuning to preserve prior knowledge. Personalization~\cite{wu2025personalized, liu2025sparse} and adaptive~\cite{yu2025efficient, zhong2025sacfl} methods address heterogeneity through task-dependent model adaptation~\cite{li2025complementary} or dynamic aggregation~\cite{nori2025federated}. pFedMxF~\cite{zhang2025pfedmxf} and SpaPFCIL~\cite{zhong2025sacfl} balance personalization and task continuity by selectively combining local adaptation with global coordination. Nevertheless, most existing FCIL methods rely on conventional naive parameter aggregation strategies, tending to overwrite the knowledge boundaries of previous classes, leading to unstable performance across phases and clients.

\subsection{MoE-based Federated Learning}
\textit{Mixture-of-Experts} (MoE)~\cite{yuksel2012twenty, zhou2022mixture, masoudnia2014mixture} has emerged as a powerful architecture to address the challenges of client heterogeneity and scalability in FL~\cite{feng2025pm, yi2024pfedmoe, zhuang2025personalized, xie2025dflmoe}, which enhances model capacity through expert modularization and task routing~\cite{mei2024fedmoe, liang2025mixture}. Nonetheless, integrating MoE into FL introduces unique challenges, such as failing to adapt to heterogeneous clients and evolving tasks~\cite{chen2025efficient}, imbalanced loading for dominant experts~\cite{chi2022representation, qiu2025demons}. To address these issues, recent works have explored MoE-based aggregation~\cite{mei2024fedmoe, zhan2024fedmoe}, which employs globally shared expert pools with the activation per client or task. Dynamic routing and enhanced gating methods~\cite{miao2025fedvla, farhat2025learning, radwan2025feddg} select experts based on gradient or reward feedback. Personalized MoE structures~\cite{zhuang2025personalized, yi2024pfedmoe, feng2025pm} provide client-specific sub-expert sets to enhance specialization. Despite these advancements, existing MoE-based methods fail to reconcile adaptability with knowledge retention in class-incremental settings and accommodate asynchronous task arrivals, motivating the need for a framework that routes knowledge dynamically without compromising previously acquired representations.

\begin{figure*}[t]
\centerline{\includegraphics[width=1.0\textwidth, trim=0 0 0 0,clip]{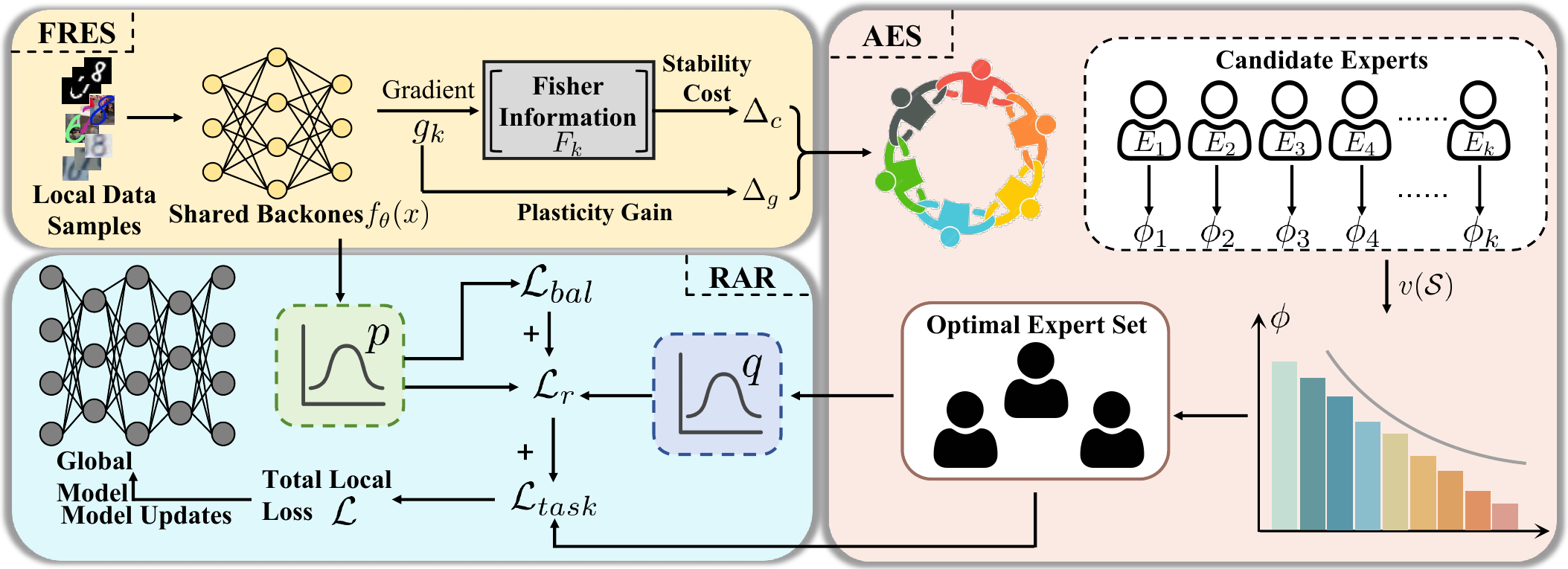}}
\caption{Overview illustration of the proposed \textsc{FedFMX} framework.}
\label{framework}
\end{figure*}

\section{Methodology}
In this section, we present the detailed methodology of \textsc{FedFMX}. We begin with the preliminaries of FCIL in \S~\ref{preliminaries} and elaborate on the FRES module design in \S~\ref{fisher_route}. We describe the AES module with the marginal contribution metric in \S~\ref{Adp_Expert_Selection}. Finally, we present the route-aware regularization in \S~\ref{regularization}. The overall framework is shown in Figure~\ref{framework}

\subsection{Preliminaries} \label{preliminaries}
We extend the standard \textit{class-incremental learning} (CIL) to federated settings. Given $N$ distributed clients that collaboratively and incrementally train a shared global model, each client $i \in \mathcal{I} = \{1, \dots, N\}$ receives a sequence of streaming tasks $\{\mathcal{T}_{i}^{(t)}\}_{t=1}^{T}$, where $\mathcal{T}_{i}^{(t)} = \{\mathcal{X}_{i}^{(t)}, \mathcal{Y}_{i}^{(t)}\}$ induces a local dataset $\mathcal{D}_{i}^{(t)} = \{(\boldsymbol{x}_{i,j}^{(t)}, \\ y_{i,j}^{(t)})\}_{j=1}^{n_{i}^{(t)}}$, where $n_{i}^{(t)} = |\mathcal{D}_{i}^{(t)}|$, the input samples $\boldsymbol{x}_{i,j}^{(t)} \in \mathcal{X}$ and the labels $y_{i,j}^{(t)} \in \mathcal{Y}_{i}^{(t)} \subseteq \mathcal{Y}$ such that the labels in different tasks are disjoint per client across stages, i.e., $\mathcal{Y}_{i}^{(t)} \cap \mathcal{Y}_{i}^{(t^{\prime})} = \varnothing$ for $t \neq t^{\prime}$. For each client $i$, when training on $\mathcal{D}_{i}^{(t)}$, the data from previous tasks $\{\mathcal{T}_{i}^{(k)}\}_{k=1}^{t-1}$ are inaccessible. Let $f(\cdot; \boldsymbol{w})$ denote the global model. The optimization objective can be characterized as: 
\begin{align}
\min_{\boldsymbol{w}} \frac{1}{D}  \sum\nolimits_{t=1}^{T} \sum\nolimits_{i=1}^{N} \sum\nolimits_{j=1}^{n_{i}^{(t)}} \mathcal{L}(f(\boldsymbol{x}_{i,j}^{(t)}; \boldsymbol{w}), y_{i,j}^{(t)}),
\end{align}
where $D = \sum\nolimits_{t=1}^{T}\sum\nolimits_{i=1}^{N} n_{i}^{(t)}$, and $\mathcal{L}$ denotes the task loss. 

\subsection{Fisher-Routed Expert Scoring Design} \label{fisher_route}
In FCIL, catastrophic forgetting and gradient interference are exacerbated by Non-IID data and asynchronous class arrivals across clients. Gradients from various tasks are entangled within a shared parameter space, limiting representational capacity. To mitigate such conflicts, we decompose the model into a shared backbone and multiple expert-specific parameter subspaces, enabling routing-conditioned optimization. Note that the expert subspaces are not statically tied to particular clients or classes; instead, they are dynamically activated at the sample level, facilitating adaptive and collaborative knowledge allocation across tasks. However, routing based solely on loss or fixed Top-$K$ gating may cause over-specialization and amplify forgetting. To address this issue, we propose a \textit{Fisher-Routed Expert Scoring} (\textbf{FRES}) module that formulates routing as a stability–plasticity trade-off. For each input sample, FRES constructs a Fisher-informed suitability score that jointly evaluates \textit{plasticity gain}, reflecting an expert’s capacity to absorb new information, and \textit{stability cost}, quantifying the risk of disrupting curvature-sensitive directions associated with prior tasks.

Concretely, we instantiate the decomposition by partitioning the parameters into a shared backbone $f_{\theta}$ and $K$ expert-specific subspaces $\{\Theta_k\}_{k=1}^{K}$, where each $\Theta_k$ comprises a dedicated body $E_{\varphi_k}$ and classification head $W_{\psi_k}$, forming independent adaptation channels rather than merely attaching personalized heads. Given any input sample $x$, expert $k$ produces logits $h_k(x)=W_{\psi_k} (E_{\varphi_k}(f_{\theta}(x)))$ with loss $\ell_k(x,y)=\mathrm{CE} (h_k(x), y)$~\cite{shazeer2017, mu2025comprehensive}. We approximate parameter importance via the \textit{empirical Fisher information matrix} $\mathbb{E}[g_k g_k^\top]$, $g_k = \nabla_{\psi_k}\ell_k(x,y)$, which captures the local curvature of each expert subspace. Unlike classical consolidation methods that apply Fisher statistics as an update regularizer~\cite{tan2026fggm, kirkpatrick2017overcoming}, we use them as a pre-update routing criterion to assess compatibility between incoming gradients and expert subspace. For efficiency, we adopt \textit{diagonal approximation} and maintain an \textit{exponential moving average} of squared gradients as:
\begin{align} \label{Fishe_update}
F_{k} \leftarrow \rho F_{k} + (1 - \rho) (g_{k} \odot g_{k} ),  g_{k} \triangleq \nabla_{\psi_{k}} \ell_{k}(x,y),
\end{align}
where $\rho \in [0,1)$ represents the decay rate and $F_{k}$ captures the influence of the parameters of expert $k$ in past predictions. Notably, we maintain diagonal Fisher statistics only for the classification heads and reuse gradients computed for backpropagation, without additional backward passes or second-order computations.

To ensure comparability, we apply the gradient normalization $\tilde{g}_{k} = g_{k}/\|g_{k}\|_{2}$ and standardize $F_{k}$ per dimension and then apply a mapping to obtain $\widetilde{F}_{k}$. Subsequently, we define two metrics to characterize expert $k$'s interaction with the sample: (\lowercase\expandafter{\romannumeral1}) \textit{Stability Cost.} The expected loss of prior knowledge is estimated:
\begin{align} \label{stable}
\Delta_{c}^{(k)} = \frac{1}{2} \|\mathrm{diag}(\widetilde{F}_{k})^{1/2} \tilde{g}_{k} \|_{2}^{2},
\end{align}
which penalizes gradient directions salient in the current update yet important historically, based on Fisher statistics accumulated before the current sample. (\lowercase\expandafter{\romannumeral2}) \textit{Plasticity Gain.} The potential improvement from learning the sample is:
\begin{align} \label{gain}
\Delta_{g}^{(k)} = \|g_{k}\|_{2}^{2} / (\mathrm{Tr}(\widetilde{F}_{k}) + \varepsilon),
\end{align}
where the constant $\varepsilon$ is for numerical stability. $\Delta_{g}^{(k)}$ approximates the effective learning gain by balancing the gradient magnitude against local parameter rigidity. Here, $\|g_k\|_2^{2}$ represents the first-order learning signal induced by the sample, while $\mathrm{Tr}(\widetilde{F}_k)$ captures the curvature and historical importance of the expert parameters. Thus, $\Delta_g^{(k)}$ favors experts that can achieve substantial loss reduction while operating in relatively flat parameter regions. Larger $\Delta_g^{(k)}$ values therefore indicate stronger and safer responsiveness to new information.

Unlike conventional dynamic head or expert selection strategies that primarily rely on instantaneous loss or gating scores, FRES calculates $\Delta_{c}^{(k)}$ and $\Delta_{g}^{(k)}$ before activation for each candidate expert during the training phase to evaluate trade-offs between learning potential and forgetting risk, enabling the routing layer to avoid over-specialized or fragile experts. 

\subsection{Adaptive Experts Selection} \label{Adp_Expert_Selection}
To enable context-aware and sample-specific expert activation, we propose an \textit{Adaptive Expert Selection} (\textbf{AES}) module. In contrast to the conventional static Top-1 or Top-$K$ gating method, which fails to adapt to evolving task demands or learning phases, AES dynamically determines the optimal subset of experts $\mathcal{K}_{b}$ for each input sample $x_{b}$ from the batch $\{(x_{b}, y_{b})\}_{b=1}^B$. The selection process explicitly considers the \textit{collaborative value} and \textit{individual contribution}, promoting balanced specialization and avoiding the overfitting to dominant experts.

We formulate the expert selection as a cooperative game over the candidate expert set $\mathcal{C} \!=\! \{1, \dots, K\}$. For any subset $\mathcal{S} \!\subseteq\! \mathcal{C}$, we define the value function $v(\mathcal{S})$ to quantify the utility of activating experts in $\mathcal{S}$ for any sample $(x, y)$:
\begin{align} \label{value_function}
v(\mathcal{S}) = - \ell_{\mathcal{S}}(x,y) - \alpha \sum\nolimits_{k \in \mathcal{S}} \Delta_{c}^{(k)} + (1 - \alpha)\overline{\Delta}_{g}(\mathcal{S}),
\end{align}
where $\alpha \in [0,1]$ governs the balance between prediction accuracy and forgetting risk; $\ell_{\mathcal{S}}(x,y) = \frac{1}{|\mathcal{S}|}\sum\nolimits_{k\in\mathcal{S}}\ell_k(x,y)$ denotes the averaged loss under expert subset $\mathcal{S}$; $\overline{\Delta}_{g}(\mathcal{S}) = \frac{1}{|\mathcal{S}|}\sum\nolimits_{k\in\mathcal{S}} \tilde{\Delta}_{g}^{(k)}$ with $\tilde{\Delta}_{g}^{(k)}$ as the normalized $\Delta_{g}^{(k)}$. (\ref{value_function}) captures the dual goals of minimizing loss and maximizing gain, while accounting for the potential risk of disrupting previously acquired knowledge. To allow incremental $\mathcal{O}(1)$ updates when expanding $\mathcal{S}_{j-1}$ to $\mathcal{S}_j$, we maintain cached aggregates $\{\sum\nolimits_{\ell}, \sum\nolimits_{\Delta_{c}}, \sum\nolimits_{\tilde{\Delta}_g}, |\mathcal{S}|\}$ with $\sum\nolimits_{\Delta_{c}}(\mathcal{S}) = \sum\nolimits_{k\in\mathcal{S}}\Delta_{c}^{(k)}$, $\sum\nolimits_{\tilde{\Delta}_{g}}(\mathcal{S}) = \sum\nolimits_{k\in\mathcal{S}}\tilde{\Delta}_{g}^{(k)}$. The individual marginal contribution of each expert $k \in \mathcal{C}$ is given by:
\begin{align} \label{shapley_value}
\phi_{k} = \sum\nolimits_{\mathcal{S} \subseteq \mathcal{C} \setminus \{k\}}  \frac{|\mathcal{S}|! (K - |\mathcal{S}| - 1)!}{K!} [v(\mathcal{S} \cup \{k\}) - v(\mathcal{S})], 
\end{align}
which quantifies the expected marginal contribution of each expert $k$ over all potent coalitions. Then, to choose the activated expert set $\mathcal{K}_{b}$ for sample $x_{b}$, we sort all experts in descending order of their marginal contribution values by $\phi_{k_{(1)}} \\ \geq \dots \geq \phi_{k_{(K)}}$. Define $\mathcal{S}_{0} = \varnothing$, we iteratively construct subsets $\mathcal{S}_{j} = \mathcal{S}_{j-1} \cup \{k_{(j)}\}$ for $j \in \mathcal{C}$, and compute the marginal coalition gain:
\begin{align}
\Delta v_{j} = v(\mathcal{S}_{j}) - v(\mathcal{S}_{j-1}),
\end{align}
which terminates at the smallest index $j^{\star} \geq 2$ satisfying $\Delta v_{j}  \leq  \mu \overline{\Delta v}_{1:j-1}$ with $\overline{\Delta v}_{1:j-1} = \frac{1}{j-1} \sum\nolimits_{p=1}^{j-1} \Delta v_{p}$, where $\mu \in [0,1]$ controls the strictness of diminishing gain. A smaller $\mu$ leads to earlier stopping and fewer experts. The final activation set for the sample $x_{b}$ is given by $\mathcal{K}_{b} = \mathcal{S}_{j^{\star}}$. 

AES module formulates expert collaboration as a cooperative game, jointly accounting for individual contribution and coalition value to avoid greedy maximization of $v(\cdot)$, thereby mitigating expert over-concentration and forgetting. Since the expert pool in FCIL generally satisfies $K\leq 8$~\cite{yi2026pFedMoE}, we compute exact Shapley values; for larger $K$, permutation-based estimators can be used instead.

\subsection{Routing-aware Regularization} \label{regularization}
To guarantee stable and fair expert utilization and address following issues:~(\lowercase\expandafter{\romannumeral1})~\textit{Expert Over-concentration:} Routing decisions concentrated on a few dominant experts, leading to resource underutilization and training instability; (\lowercase\expandafter{\romannumeral2}) \textit{Stage-End Forgetting Risk:} During later incremental stages, the risk of forgetting learned classes remains high, we adopt the routing-aware regularization (\textbf{RAR}) strategy with a load-balancing regularization term and distilled gating mechanism, which extend the stability–plasticity trade-off from expert selection to optimization.

We first introduce a load-balancing regularization term to encourage smoother expert activation distributions and prevent overloading. Specifically, for a training batch of size $B$, we define the entropy-based regularizer as:
\begin{align} \label{loss_bal}
\mathcal{L}_{bal} = \frac{1}{B} \sum\nolimits_{b=1}^{B} \sum\nolimits_{k=1}^{K} p(k \mid x_{b}) \log p(k \mid x_{b}),
\end{align}
where $p(k \mid x_{b})$ indicates the soft expert routing distribution for sample $x_{b}$. Minimizing the term $\mathcal{L}_{bal}$ discourages sharp routing distributions and mitigates the overload of minority experts. Then, to decouple routing from marginal contribution calculation while retaining the training-time benefit, we adopt a distillation-based regularization scheme. Specifically, we convert the expert importance scores $\phi_k$ in (\ref{shapley_value}) into a soft \textit{teacher} distribution $q(k \mid x_{b}) = \mathrm{softmax}(\phi_{k} / \tau_{r})$ with temperature hyperparameter $\tau_{r} > 0$ controlling the sharpness. Subsequently, we instantiate a lightweight gating network $g_{\eta}$ that maps the backbone feature to logits $z^{(g)}\in\mathbb{R}^{K}$, which are converted into the \textit{student} distribution as $p(k\mid x_{b})=\mathrm{softmax}(z^{(g)}_{k}/T)$ with distillation temperature $T > 0$. To ensure that the student network captures the expert activation patterns implied by the marginal contribution scores, the distillation loss is defined as:
\begin{align} \label{loss_route}
\mathcal{L}_{r} = \frac{1}{B} \sum\nolimits_{b=1}^{B} \mathrm{KL} ( q(\cdot \mid x_{b}) \| p(\cdot \mid x_{b}) ),
\end{align}
which regularizes the gating network to align with the semantics of Shapley-based expert selection while allowing efficient routing at inference. This distillation objective serves two key purposes: (\textit{i}) it eliminates the need to compute costly marginal contributions during deployment, and (\textit{ii}) it encourages the learned routing mechanism to generalize expert specialization patterns beyond the training distribution. By combining the above components into a unified loss function for local training. For each batch $\{(x_{b}, y_{b})\}_{b=1}^B$, the total local loss is:
\begin{align} \label{total_loss}
\mathcal{L} = \mathcal{L}_{task} + \beta  \mathcal{L}_{bal} + \gamma \mathcal{L}_{r}, 
\end{align}
where $\mathcal{L}_{task} = \frac{1}{B} \sum\nolimits_{b=1}^{B} \sum\nolimits_{k \in \mathcal{K}_{b}} w_{k} \ell_{k}$; $\beta$ and $\gamma$ control the strength of load balancing and route distillation. For training stability, we define the weight $w_{k} = \frac{\max(\phi_{k}, 0)}{\sum\nolimits_{j \in \mathcal{K}_{b}} \max(\phi_j, 0)}$ to balance expert updates within the active set, allowing experts with strong positive marginal values to contribute, while suppressing noisy or redundant experts. We provide the detailed algorithm in Appendix~A.

\subsection{Discussion}
\textsc{FedFMX} enables stability-aware training and efficient deployment via adaptive expert routing and distillation-based gating. During training, marginal contribution scores are transformed into a teacher distribution $q(\cdot \mid x_b)$, based on which AES selects a hard expert subset $\mathcal{K}_b$ for forward and backward propagation. A lightweight gating network produces a student distribution $p(\cdot \mid x_b)$, trained to match $q(\cdot \mid x_b)$ via KL distillation. The marginal contribution scores are used exclusively for expert subset selection and contribution-based weighting, and are treated as non-differentiable routing signals that do not participate in gradient updates. During inference, Top-$1$/Top-$K$ experts are selected according to the distilled distribution $p(\cdot \mid x)$ and aggregated with normalized gating weights, enabling forward-only inference without marginal contribution computation.

\section{Theoretical Analysis}

We analyze convergence under standard federated optimization assumptions, including Lipschitz smoothness, unbiased gradients, and bounded first- and second-moment conditions (see Appendix~B). AES induces an iteration-dependent projection onto activated expert subspaces, resulting in a projected-gradient FedAvg update. The shared backbone is updated using all samples, while the projection applies only to expert-specific parameters. Although FRES determines the projection and regularizers modify the local objective, the overall update remains projected-gradient optimization. Following~\cite{stich2018sparsified, karimireddy2019error}, we establish convergence under bounded residual gradient energy outside the activated subspace. Proofs are provided in Appendix~C-E.

\begin{assumption} \label{assumption_grad_residual}
(Bounded Activated-Subspace Gradient Residual) The AES module induces an activation-dependent parameter subspace $\mathcal{S}_{i}(\boldsymbol{w}) \subseteq \mathbb{R}^d$ with projection operator $P_{\mathcal{S}_{i}(\boldsymbol{w})}$. The effective update under AES corresponds to the projected gradient $\hat{g}_{i}(\boldsymbol{w}) = P_{\mathcal{S}_{i}(\boldsymbol{w})} g_{i}(\boldsymbol{w})$. Assume that the discarded gradient energy outside the activated subspace is uniformly bounded, i.e., there exists $\delta \geq 0$ such~that
\begin{align}
\mathbb{E}[\| (I - P_{\mathcal{S}_{i}(\boldsymbol{w})}) g_{i}(\boldsymbol{w})\|^{2}] \leq \delta^{2}, \forall i, \boldsymbol{w}.
\end{align}
\end{assumption}
Based on the assumptions, we can derive the following Lemma and Theorem.
\begin{lemma}\label{lemma_one_step_descent}
\textbf{Local Training.} Consider the local update on client $i \in \{1, \ldots, N\}$ by $\boldsymbol{w}^{+} = \boldsymbol{w} - \eta \hat{g}_{i}(\boldsymbol{w})$, for $\eta \leq \frac{1}{2\beta}$, the expected convergent upper bound of the local objective function for any consecutive global iterations is given by
\begin{align} \label{one_step_descent_bound}
\mathbb{E} [F_{i}(\boldsymbol{w}^{+}) \mid \boldsymbol{w}] \leq F_{i}(\boldsymbol{w}) - \frac{\eta}{4}\|\nabla F_{i}(\boldsymbol{w})\|^{2} + 2\sigma^{2}\beta\eta^{2} + (\beta\eta^{2} + \frac{\eta}{2})\delta^{2}.
\end{align}
\end{lemma}

\begin{lemma}\label{lemma_local_drift}
\textbf{Local Drift.} Consider $E$ local steps. Define $\boldsymbol{w}_{i}^{t,0}=\boldsymbol{w}^{t}$, and $\boldsymbol{w}_{i}^{t,e+1}=\boldsymbol{w}_{i}^{t,e}-\eta \hat{g}_{i}(\boldsymbol{w}_{i}^{t,e})$, for $e= \{0,\ldots,E-1\}$ and $\eta \leq \frac{1}{4 \beta E}$, then
\begin{align}\label{local_drift}
\mathbb{E} [\|\boldsymbol{w}_{i}^{t,E}-\boldsymbol{w}^{t}\|^{2} \mid \boldsymbol{w}^{t}] \leq 16 \eta^{2} E^{2} (\|\nabla F_i(\boldsymbol{w}^{t})\|^{2} + \sigma^{2} + \delta^{2}).
\end{align}
\end{lemma}

\begin{theorem}\label{theorem_global_convergence}
\textbf{Global Convergence.} Let $F(\boldsymbol{w}) \triangleq \sum_{i=1}^{N} \lambda_{i} F_{i}(\boldsymbol{w})$ with $\sum_{i=1}^{N} \lambda_{i}=1$. For $\eta \leq \min \{\frac{1}{2\beta}, \frac{1}{4 \beta E}, \frac{\kappa}{8 \beta E \kappa_{G}^{2}}, \frac{\kappa^{2}}{48 E M_{V}}\}$ and any $T \geq 1$, it satifies
\begin{align} \label{convergence_bound}
\frac{1}{T}\sum_{t=0}^{T-1}\mathbb{E} [\|\nabla F(\boldsymbol{w}^{t})\|^{2}] \!\leq\! \frac{8 (F(\boldsymbol{w}^{0}) \!-\! F(\boldsymbol{w}^{\star}))}{\kappa \eta E T} + 6(2M \!+\! 3(\sigma^{2}+\delta^{2}))(\frac{2}{\kappa^{2}} \!+\! \frac{1}{\kappa}).
\end{align}
\end{theorem}

In Theorem~\ref{theorem_global_convergence}, the first term decays as $\mathcal{O}(1/T)$, while the second term forms an error floor determined by the gradient noise and the activated-subspace residual $\delta$. In the deterministic and full-subspace case, the bound reduces to $\mathcal{O}(1/T)$.

\section{Numerical Experiments}
\subsection{Experiment Setups}

\textbf{Datasets \& Model.}\ We evaluate our method on four real-world image classification tasks: CIFAR-10/CIFAR-100~\cite{krizhevsky2009learning}, Tiny-ImageNet~\cite{le2015tiny}, and DomainNet~\cite{peng2019moment}. We follow the experimental settings in~\cite{zhang2025pfedmxf, qi2025classwise} and adopt two backbone architectures for training: ResNet-18~\cite{he2016deep} and ViT-B/16~\cite{dosovitskiy2020image} (in Appendix~G), where the ViT-B/16 model is initialized with self-supervised pretrained weights from DINO~\cite{caron2021emerging}, which is widely used in CIL. 

\begin{table*}[t]
\centering
\caption{Accuracy comparison of \setlength{\fboxsep}{1pt}\colorbox{tablecolor}{\textsc{FedFMX}} and other benchmark methods on ResNet-18 backbone under the fine-grained setting. The best accuracy is in \textbf{bold}.}
\renewcommand\arraystretch{1.0}
\label{tab:overall}
\vspace{-18pt}
\begin{subtable}{\linewidth}
\centering
\caption{Under the QLI setting}
\vspace{-8pt}
\resizebox{1.0\textwidth}{!}{\begin{tabular}
{c|ccc|ccc|ccc|ccc} 
\toprule[1.2pt]

\multirow{2}{*}{\multirowcell{2}{\centering\textbf{Method}}} & \multicolumn{3}{c|}{\centering\textbf{CIFAR-10}} & \multicolumn{3}{c|}{\centering\textbf{CIFAR-100}} & \multicolumn{3}{c|}{\centering\textbf{Tiny-ImageNet}} & \multicolumn{3}{c}{\centering\textbf{DomainNet}}  \\ \cmidrule[0.5pt](l{1pt}r{0pt}){2-13}

& $\alpha_{Q} = 2$ & $\alpha_{Q} = 4$ & $\alpha_{Q} = 6$ & $\alpha_{Q} = 2$ & $\alpha_{Q} = 4$ & $\alpha_{Q} = 6$ & $\alpha_{Q} = 2$ & $\alpha_{Q} = 4$ & $\alpha_{Q} = 6$ & $\alpha_{Q} = 2$ & $\alpha_{Q} = 4$ & $\alpha_{Q} = 6$ \\ \cmidrule[0.8pt](l{1pt}r{0pt}){1-13}

\textsc{FedMut} & 18.52$\pm \scriptstyle{2.2}$ & 23.41$\pm \scriptstyle{2.6}$ & 28.96$\pm \scriptstyle{2.7}$ &  5.43$\pm \scriptstyle{0.9}$ &  7.76$\pm \scriptstyle{1.4}$ &  9.85$\pm \scriptstyle{1.8}$ &  2.96$\pm \scriptstyle{0.8}$ &  4.10$\pm \scriptstyle{0.6}$ &  5.21$\pm \scriptstyle{0.7}$ & 4.72$\pm \scriptstyle{0.9}$ & 6.84$\pm \scriptstyle{1.2}$ & 8.93$\pm \scriptstyle{1.1}$  \\  

\textsc{FedCross} & 20.07$\pm \scriptstyle{2.1}$ & 25.48$\pm \scriptstyle{2.3}$ & 31.03$\pm \scriptstyle{2.2}$ &  6.79$\pm \scriptstyle{1.3}$ & 9.18$\pm \scriptstyle{1.6}$ & 11.24$\pm \scriptstyle{1.7}$ & 3.41$\pm \scriptstyle{0.6}$ & 4.66$\pm \scriptstyle{0.7}$ & 5.84$\pm \scriptstyle{0.6}$ & 5.61$\pm\scriptstyle{1.1}$ & 7.93$\pm\scriptstyle{1.4}$ & 10.24$\pm\scriptstyle{1.2}$  \\  

\textsc{FedLSA} & 16.44$\pm \scriptstyle{1.9}$ & 21.36$\pm \scriptstyle{2.6}$ & 26.62$\pm \scriptstyle{2.4}$ &  4.84$\pm \scriptstyle{0.8}$ &  6.86$\pm \scriptstyle{1.2}$ &  8.91$\pm \scriptstyle{1.5}$ &  2.83$\pm \scriptstyle{0.6}$ &  3.81$\pm \scriptstyle{0.9}$ &  4.86$\pm \scriptstyle{0.7}$ & 4.18$\pm\scriptstyle{0.8}$ & 6.11$\pm\scriptstyle{1.3}$ & 8.21$\pm\scriptstyle{1.0}$  \\  

\textsc{FedWMSAM} & 19.61$\pm \scriptstyle{2.3}$ & 24.62$\pm \scriptstyle{2.1}$ & 29.57$\pm \scriptstyle{2.8}$ &  6.52$\pm \scriptstyle{1.9}$ &  8.85$\pm \scriptstyle{1.5}$ & 10.83$\pm \scriptstyle{1.6}$ &  3.24$\pm \scriptstyle{0.6}$ &  4.46$\pm \scriptstyle{0.6}$ &  5.62$\pm \scriptstyle{0.7}$ & 5.23$\pm\scriptstyle{0.9}$ & 7.58$\pm\scriptstyle{1.1}$ & 9.82$\pm\scriptstyle{1.6}$  \\  

\textsc{pFedMoE} & 41.47$\pm \scriptstyle{2.2}$ & 47.96$\pm \scriptstyle{2.1}$ & 53.56$\pm \scriptstyle{2.5}$ & 24.83$\pm \scriptstyle{1.4}$ & 30.76$\pm \scriptstyle{1.3}$ & 34.82$\pm \scriptstyle{1.1}$ &  8.79$\pm \scriptstyle{0.5}$ & 11.02$\pm \scriptstyle{0.8}$ & 12.18$\pm \scriptstyle{0.7}$  & 21.76$\pm\scriptstyle{1.4}$ & 27.92$\pm\scriptstyle{1.2}$ & 32.18$\pm\scriptstyle{1.3}$ \\  

\textsc{FedEWC} & 17.11$\pm \scriptstyle{2.5}$ & 21.14$\pm \scriptstyle{2.4}$ & 30.52$\pm \scriptstyle{2.6}$ & 9.14$\pm \scriptstyle{1.4}$ & 12.32$\pm \scriptstyle{1.0}$ & 13.46$\pm \scriptstyle{1.5}$ & 4.43$\pm \scriptstyle{0.7}$ & 5.07$\pm \scriptstyle{0.8}$ & 6.18$\pm \scriptstyle{0.9}$ & 6.98$\pm\scriptstyle{1.1}$ & 9.11$\pm\scriptstyle{1.3}$ & 11.37$\pm\scriptstyle{1.5}$  \\  

\textsc{FedLwF} & 35.93$\pm \scriptstyle{2.9}$ & 39.25$\pm \scriptstyle{2.1}$ & 52.47$\pm \scriptstyle{2.9}$ & 20.12$\pm \scriptstyle{1.1}$ & 26.09$\pm \scriptstyle{1.2}$ & 29.34$\pm \scriptstyle{1.3}$ & 4.85$\pm \scriptstyle{0.9}$ & 7.14$\pm \scriptstyle{0.6}$ & 11.22$\pm \scriptstyle{0.7}$ & 18.42$\pm \scriptstyle{1.3}$ & 23.96$\pm \scriptstyle{1.5}$ & 27.85$\pm \scriptstyle{1.2}$  \\  

\textsc{TARGET} & 33.29$\pm \scriptstyle{2.4}$ & 36.67$\pm \scriptstyle{2.1}$ &49.75$\pm \scriptstyle{2.3}$ & 15.29$\pm \scriptstyle{1.1}$ & 22.27$\pm \scriptstyle{0.7}$ & 26.28$\pm \scriptstyle{1.2}$ & 6.16$\pm \scriptstyle{0.8}$ & 6.48$\pm \scriptstyle{0.5}$ & 8.45$\pm \scriptstyle{1.1}$ & 13.52$\pm \scriptstyle{1.0}$ & 19.83$\pm \scriptstyle{1.8}$ & 24.26$\pm \scriptstyle{1.4}$ \\  

\textsc{LANDER} & 36.68$\pm \scriptstyle{2.5}$ & 40.49$\pm \scriptstyle{2.4}$ & 59.33$\pm \scriptstyle{2.4}$ & 28.28$\pm \scriptstyle{1.6}$ & 35.11$\pm \scriptstyle{1.1}$ & 38.45$\pm \scriptstyle{1.6}$ & 9.22$\pm \scriptstyle{0.9}$ & 9.53$\pm \scriptstyle{0.7}$ & 11.23$\pm \scriptstyle{0.8}$ & 24.17$\pm \scriptstyle{1.6}$ & 30.85$\pm \scriptstyle{1.3}$ & 34.94$\pm \scriptstyle{1.1}$  \\ 

\textsc{Re-Fed} & 51.96$\pm \scriptstyle{3.1}$ & 55.13$\pm \scriptstyle{2.9}$ & 59.58$\pm \scriptstyle{2.5}$ & 30.55$\pm \scriptstyle{1.5}$ & 40.28$\pm \scriptstyle{0.9}$ & 41.52$\pm \scriptstyle{1.5}$ & 9.97$\pm \scriptstyle{0.6}$ & 12.06$\pm \scriptstyle{0.6}$ & 13.37$\pm \scriptstyle{0.7}$ & 26.94$\pm \scriptstyle{1.7}$ & 36.18$\pm \scriptstyle{1.3}$ & 38.42$\pm \scriptstyle{1.6}$ \\ 

\textsc{FL-CLIP} & 50.25$\pm \scriptstyle{2.9}$ & 53.26$\pm \scriptstyle{2.7}$ & 59.47$\pm \scriptstyle{2.4}$ & 29.76$\pm \scriptstyle{1.6}$ & 39.15$\pm \scriptstyle{1.2}$ & 40.87$\pm \scriptstyle{1.4}$ & 9.51$\pm \scriptstyle{0.8}$ & 11.85$\pm \scriptstyle{1.0}$ & 13.15$\pm \scriptstyle{0.6}$ & 25.83$\pm \scriptstyle{1.1}$ & 34.77$\pm \scriptstyle{1.6}$ & 37.15$\pm \scriptstyle{1.3}$ \\ 

\textsc{FedCBDR} & 57.62$\pm \scriptstyle{2.6}$ & 61.45$\pm \scriptstyle{2.0}$ & 65.79$\pm \scriptstyle{1.9}$ & 45.54$\pm \scriptstyle{1.4}$ & 46.97$\pm \scriptstyle{1.2}$ & 48.15$\pm \scriptstyle{1.3}$ & 13.44$\pm \scriptstyle{0.7}$ & 15.32$\pm \scriptstyle{0.6}$ & 16.69$\pm \scriptstyle{0.9}$ & 41.26$\pm \scriptstyle{1.8}$ & 44.37$\pm \scriptstyle{1.2}$ & 46.52$\pm \scriptstyle{1.4}$ \\ 

\textsc{pFedMxF} & 56.99$\pm \scriptstyle{2.8}$ & 61.38$\pm \scriptstyle{2.1}$ & 65.56$\pm \scriptstyle{2.1}$ & 45.07$\pm \scriptstyle{1.5}$ & 46.88$\pm \scriptstyle{1.4}$ & 47.78$\pm \scriptstyle{1.5}$ & 13.05$\pm \scriptstyle{0.7}$ & 15.27$\pm \scriptstyle{0.7}$ & 16.58$\pm \scriptstyle{0.8}$ & 40.74$\pm \scriptstyle{1.3}$ & 43.92$\pm \scriptstyle{1.7}$ & 45.98$\pm \scriptstyle{1.5}$ \\ \midrule[0.8pt]

\cellcolor{tablecolor}\textsc{FedFMX}(Ours) & \cellcolor{tablecolor}\textbf{60.35}$\pm \scriptstyle{2.6}$ & \cellcolor{tablecolor}\textbf{65.28}$\pm \scriptstyle{1.8}$ & \cellcolor{tablecolor}\textbf{68.94}$\pm \scriptstyle{2.1}$ & \cellcolor{tablecolor}\textbf{49.87}$\pm \scriptstyle{1.2}$ & \cellcolor{tablecolor}\textbf{51.18}$\pm \scriptstyle{1.3}$ & \cellcolor{tablecolor}\textbf{53.24}$\pm \scriptstyle{1.2}$ & \cellcolor{tablecolor}\textbf{16.76}$\pm \scriptstyle{0.8}$ & \cellcolor{tablecolor}\textbf{19.47}$\pm \scriptstyle{0.5}$ & \cellcolor{tablecolor}\textbf{21.36}$\pm \scriptstyle{0.6}$ & \cellcolor{tablecolor}\textbf{44.82}$\pm \scriptstyle{1.5}$ & \cellcolor{tablecolor}\textbf{47.63}$\pm \scriptstyle{1.1}$ & \cellcolor{tablecolor}\textbf{49.76}$\pm \scriptstyle{1.2}$  \\ 

\bottomrule[1.2pt]
\end{tabular}}
\label{accuracy_QLI}
\end{subtable}

\vspace{5pt}  

\begin{subtable}{\linewidth}
\centering
\caption{Under the DIL setting}
\vspace{-8pt}
\resizebox{1.0\textwidth}{!}{\begin{tabular}
{c|ccc|ccc|ccc|ccc} 
\toprule[1.2pt]
\multirow{2}{*}{\multirowcell{2}{\centering\textbf{Method}}} & \multicolumn{3}{c|}{\centering\textbf{CIFAR-10}} & \multicolumn{3}{c|}{\centering\textbf{CIFAR-100}} & \multicolumn{3}{c|}{\centering\textbf{Tiny-ImageNet}} & \multicolumn{3}{c}{\centering\textbf{DomainNet}}  \\ \cmidrule[0.5pt](l{1pt}r{0pt}){2-13}

& $\alpha_{D} = 0.1$ & $\alpha_{D} = 0.5$ & $\alpha_{D} = 1.0$ & $\alpha_{D} = 0.1$ & $\alpha_{D} = 0.5$ & $\alpha_{D} = 1.0$ & $\alpha_{D} = 0.1$ & $\alpha_{D} = 0.5$ & $\alpha_{D} = 1.0$ & $\alpha_{D} = 0.1$ & $\alpha_{D} = 0.5$ & $\alpha_{D} = 1.0$ \\ \cmidrule[0.8pt](l{1pt}r{0pt}){1-13}

\textsc{FedMut} & 15.62$\pm \scriptstyle{2.3}$ & 20.11$\pm \scriptstyle{2.1}$ & 26.48$\pm \scriptstyle{2.7}$ &  4.21$\pm \scriptstyle{1.3}$ &  6.37$\pm \scriptstyle{1.2}$ &  8.54$\pm \scriptstyle{1.0}$ &  2.41$\pm \scriptstyle{0.8}$ &  3.58$\pm \scriptstyle{0.6}$ & 4.96$\pm \scriptstyle{0.7}$ & 3.80$\pm \scriptstyle{1.4}$ & 5.61$\pm \scriptstyle{1.3}$ & 7.72$\pm \scriptstyle{1.0}$  \\  

\textsc{FedCross} & 17.38$\pm \scriptstyle{2.6}$ & 22.74$\pm \scriptstyle{2.1}$ & 28.93$\pm \scriptstyle{2.2}$ &  5.03$\pm \scriptstyle{0.9}$ &  7.41$\pm \scriptstyle{1.3}$ &  9.86$\pm \scriptstyle{1.4}$ &  2.86$\pm \scriptstyle{0.6}$ &  4.21$\pm \scriptstyle{0.7}$ &  5.64$\pm \scriptstyle{0.7}$ & 4.46$\pm \scriptstyle{1.2}$ & 6.52$\pm \scriptstyle{0.9}$ & 8.68$\pm \scriptstyle{1.1}$  \\  

\textsc{FedLSA} & 13.47$\pm \scriptstyle{2.0}$ & 18.23$\pm \scriptstyle{2.0}$ & 23.94$\pm \scriptstyle{2.1}$ & 3.56$\pm \scriptstyle{1.1}$ &  5.48$\pm \scriptstyle{0.9}$ &  7.63$\pm \scriptstyle{1.1}$ &  2.02$\pm \scriptstyle{0.5}$ &  3.01$\pm \scriptstyle{0.6}$ &  4.27$\pm \scriptstyle{0.8}$ & 3.13$\pm \scriptstyle{0.8}$ & 4.82$\pm \scriptstyle{0.9}$ & 6.71$\pm \scriptstyle{1.0}$  \\  

\textsc{FedWMSAM} & 16.82$\pm \scriptstyle{1.8}$ & 21.93$\pm \scriptstyle{2.1}$ & 27.41$\pm \scriptstyle{2.2}$ &  4.76$\pm \scriptstyle{1.8}$ &  6.89$\pm \scriptstyle{0.9}$ &  9.12$\pm \scriptstyle{1.5}$ &  2.63$\pm \scriptstyle{0.9}$ &  3.89$\pm \scriptstyle{0.6}$ &  5.31$\pm \scriptstyle{0.7}$ & 4.19$\pm \scriptstyle{1.2}$ & 6.06$\pm \scriptstyle{1.4}$ & 8.03$\pm \scriptstyle{1.1}$ \\  

\textsc{pFedMoE} & 39.84$\pm \scriptstyle{2.4}$ & 46.27$\pm \scriptstyle{1.9}$ & 52.18$\pm \scriptstyle{2.1}$ & 22.61$\pm \scriptstyle{1.4}$ & 28.73$\pm \scriptstyle{1.3}$ & 33.24$\pm \scriptstyle{1.4}$ &  7.92$\pm \scriptstyle{0.8}$ & 10.84$\pm \scriptstyle{0.8}$ & 12.63$\pm \scriptstyle{0.9}$ & 19.92$\pm \scriptstyle{1.4}$ & 25.28$\pm \scriptstyle{1.8}$ & 29.25$\pm \scriptstyle{1.3}$  \\  

\textsc{FedEWC} & 16.24$\pm \scriptstyle{2.8}$ & 20.07$\pm \scriptstyle{2.6}$ & 29.44$\pm \scriptstyle{2.3}$ & 8.22$\pm \scriptstyle{1.0}$ & 11.03$\pm \scriptstyle{0.8}$ & 12.31$\pm \scriptstyle{1.6}$ & 3.62$\pm \scriptstyle{0.5}$ & 4.39$\pm \scriptstyle{0.6}$ & 5.22$\pm \scriptstyle{0.8}$ & 7.23$\pm \scriptstyle{1.7}$ & 9.71$\pm \scriptstyle{1.3}$ & 10.83$\pm \scriptstyle{1.6}$ \\  

\textsc{FedLwF} & 34.82$\pm \scriptstyle{3.1}$ & 38.98$\pm \scriptstyle{2.3}$ & 51.53$\pm \scriptstyle{2.8}$ & 18.96$\pm \scriptstyle{1.1}$ & 25.21$\pm \scriptstyle{0.7}$ & 28.97$\pm \scriptstyle{1.2}$ & 3.97$\pm \scriptstyle{0.4}$ & 6.25$\pm \scriptstyle{0.9}$ & 10.34$\pm \scriptstyle{1.1}$ & 16.68$\pm \scriptstyle{1.3}$ & 22.18$\pm \scriptstyle{1.9}$ & 25.49$\pm \scriptstyle{1.4}$  \\  
									
\textsc{TARGET} & 32.21$\pm \scriptstyle{2.2}$ & 35.76$\pm \scriptstyle{1.7}$ & 48.62$\pm \scriptstyle{1.8}$ & 14.04$\pm \scriptstyle{0.9}$ & 21.38$\pm \scriptstyle{0.7}$ & 25.35$\pm \scriptstyle{1.3}$ & 5.21$\pm \scriptstyle{0.6}$ & 5.67$\pm \scriptstyle{0.8}$ & 7.86$\pm \scriptstyle{0.7}$ & 12.36$\pm \scriptstyle{1.5}$ & 18.81$\pm \scriptstyle{0.9}$ & 22.31$\pm \scriptstyle{1.2}$ \\  

\textsc{LANDER} & 35.73$\pm \scriptstyle{2.8}$ & 39.57$\pm \scriptstyle{2.3}$ & 58.19$\pm \scriptstyle{2.5}$ & 27.35$\pm \scriptstyle{1.8}$ & 34.02$\pm \scriptstyle{1.5}$ & 37.66$\pm \scriptstyle{1.8}$ & 8.78$\pm \scriptstyle{0.7}$ & 8.66$\pm \scriptstyle{0.5}$ & 10.24$\pm \scriptstyle{0.6}$ & 24.07$\pm \scriptstyle{1.7}$ & 29.94$\pm \scriptstyle{1.2}$ & 33.14$\pm \scriptstyle{1.4}$ \\ 

\textsc{Re-Fed} & 51.02$\pm \scriptstyle{3.4}$ & 54.21$\pm \scriptstyle{3.2}$ & 58.36$\pm \scriptstyle{2.3}$ & 29.42$\pm \scriptstyle{1.4}$ & 39.76$\pm \scriptstyle{1.2}$ & 40.87$\pm \scriptstyle{1.2}$ & 9.18$\pm \scriptstyle{0.8}$ & 11.43$\pm \scriptstyle{0.7}$ & 12.28$\pm \scriptstyle{0.9}$ & 25.89$\pm \scriptstyle{1.5}$ & 34.99$\pm \scriptstyle{1.7}$ & 35.97$\pm \scriptstyle{1.1}$ \\ 

\textsc{FL-CLIP} & 49.37$\pm \scriptstyle{3.2}$ & 52.19$\pm \scriptstyle{2.8}$ & 57.85$\pm \scriptstyle{2.4}$ & 28.59$\pm \scriptstyle{1.5}$ & 37.95$\pm \scriptstyle{1.4}$ & 39.90$\pm \scriptstyle{1.4}$ & 8.94$\pm \scriptstyle{0.7}$ & 10.91$\pm \scriptstyle{0.8}$ & 12.34$\pm \scriptstyle{1.0}$ & 25.16$\pm \scriptstyle{1.4}$ & 33.40$\pm \scriptstyle{1.2}$ & 35.11$\pm \scriptstyle{1.3}$ \\ 

\textsc{FedCBDR} & 56.49$\pm \scriptstyle{2.4}$ & 60.58$\pm \scriptstyle{1.6}$ & 64.96$\pm \scriptstyle{1.9}$ & 44.63$\pm \scriptstyle{1.3}$ & 46.27$\pm \scriptstyle{1.4}$ & 47.42$\pm \scriptstyle{1.3}$ & 12.29$\pm \scriptstyle{0.6}$ & 14.38$\pm \scriptstyle{0.6}$ & 15.72$\pm \scriptstyle{0.7}$ & 39.27$\pm \scriptstyle{1.1}$ & 42.72$\pm \scriptstyle{1.5}$ & 43.73$\pm \scriptstyle{1.6}$ \\ 

\textsc{pFedMxF} & 55.78$\pm \scriptstyle{2.6}$ & 60.43$\pm \scriptstyle{1.5}$ &  64.82$\pm \scriptstyle{1.9}$ & 43.97$\pm \scriptstyle{1.4}$ & 46.09$\pm \scriptstyle{1.6}$ & 47.12$\pm \scriptstyle{1.3}$ & 12.17$\pm \scriptstyle{0.5}$ & 14.25$\pm \scriptstyle{0.7}$ & 15.46$\pm \scriptstyle{0.6}$ & 38.69$\pm \scriptstyle{1.2}$ & 40.56$\pm \scriptstyle{1.0}$ & 41.47$\pm \scriptstyle{1.1}$ \\ \midrule[0.8pt]

\cellcolor{tablecolor}\textsc{FedFMX}(Ours) & \cellcolor{tablecolor}\textbf{59.36}$\pm \scriptstyle{2.1}$ & \cellcolor{tablecolor}\textbf{64.68}$\pm \scriptstyle{1.2}$ & \cellcolor{tablecolor}\textbf{67.46}$\pm \scriptstyle{1.5}$ & \cellcolor{tablecolor}\textbf{47.87}$\pm \scriptstyle{1.1}$ & \cellcolor{tablecolor}\textbf{50.32}$\pm \scriptstyle{1.5}$ & \cellcolor{tablecolor}\textbf{50.76}$\pm \scriptstyle{1.4}$ & \cellcolor{tablecolor}\textbf{15.88}$\pm \scriptstyle{0.5}$ & \cellcolor{tablecolor}\textbf{19.14}$\pm \scriptstyle{0.7}$ & \cellcolor{tablecolor}\textbf{22.95}$\pm \scriptstyle{0.4}$ & \cellcolor{tablecolor}\textbf{43.75$\pm \scriptstyle{0.9}$} & \cellcolor{tablecolor}\textbf{44.29$\pm \scriptstyle{1.1}$} & \cellcolor{tablecolor}\textbf{46.83$\pm \scriptstyle{1.05}$} \\ 

\bottomrule[1.2pt]
\end{tabular}}
\label{accuracy_DLI}
\end{subtable}
\vspace{-15pt}
\end{table*}

\noindent \textbf{Baselines.}\ We compare our method with following three types of baselines: (\lowercase\expandafter{\romannumeral1}) General FL methods: \textsc{FedMut}~\cite{hu2024fedmut}, \textsc{FedCross}~\cite{hu2024fedcross}, \textsc{FedLSA}~\cite{fu2025beyond}, \textsc{FedWMSAM}~\cite{li2025fedwmsam},  \textsc{pFedMoE}~\cite{yi2026pFedMoE}; (\lowercase\expandafter{\romannumeral2}) Traditional CIL methods under federated settings: \textsc{FedEWC}~\cite{kirkpatrick2017overcoming}, \textsc{FedLwF}~\cite{li2017learning}; (\lowercase\expandafter{\romannumeral3})  FCIL methods: \textsc{TARGET}~\cite{zhang2023target}, \textsc{LANDER}~\cite{tran2024text}, \textsc{Re-Fed}~\cite{li2024towards}, FL-CLIP~\cite{tan2024fl}, \textsc{FedCBDR}~\cite{qi2025classwise}, \textsc{pFedMxF}~\cite{zhang2025pfedmxf}.

\noindent \textbf{Implementation Details.}\ We employ SGD as the optimizer, a learning rate of $0.01$, momentum of $0.9$, and weight decay of $1e^{-5}$. For all datasets, we set the batch size to $64$ and the local epoch to $5$. We adopt two different types of Non-IID settings~\cite{li2022federated}: Quantity-based Label Imbalance (QLI) and Distribution-based Label Imbalance (DLI), denoted by $\alpha_{Q}$ and $\alpha_{D}$ respectively, where BLI allocates the data samples via the Dirichlet distribution. To assess robustness under varying task granularity, we consider the following two incremental settings: \textit{coarse-grained} (CIFAR-10 divided into 3 tasks, CIFAR-100 into 5 tasks, DomainNet into 5 tasks, and Tiny-ImageNet into 10 tasks); \textit{fine-grained} (CIFAR-10 divided into 5 tasks, CIFAR-100 into 10 tasks, DomainNet into 10 tasks, and Tiny-ImageNet into 20 tasks). The weight $\rho$ for (\ref{Fishe_update}) is set to $0.95$, and $\alpha=0.6$ for (\ref{value_function}). The regularization terms in (\ref{total_loss}) is weighted by $\beta = 0.05$ and $\gamma = 0.6$. Unless otherwise specified, we set $\alpha_{D}=0.5$ by default. We set $N=100$ clients, where $20\%$ of clients are randomly sampled at each round. We default the number of experts to $K=4$, where each expert shares the same architecture but maintains independent parameters, consisting of an expert body and a classification head built on top of the shared backbone. During inference, we select the Top-$2$ experts according to the routing scores. We conduct experiments on Ubuntu 22.04.4; GPU: 2NVIDIA GeForce RTX 4090. More experiment details are provided in Appendix~F.

\begin{figure*}[t]
\setlength{\abovecaptionskip}{2pt} 
    \centering
    \begin{minipage}{345pt}
        \begin{minipage}{345pt}
            \subfloat{
                \includegraphics[width=0.24\textwidth, trim= 5 5 5 5,clip]{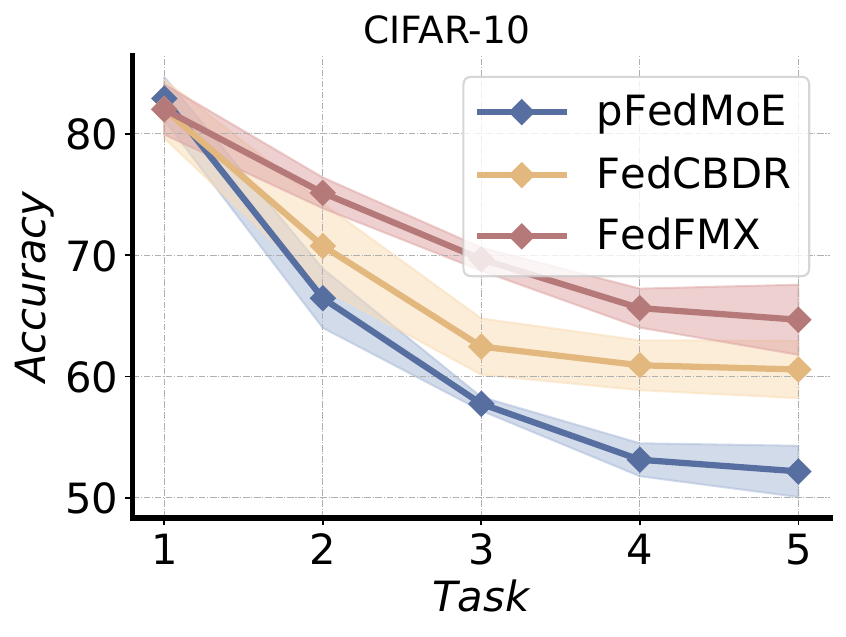}}
            \subfloat{
                \includegraphics[width=0.24\textwidth, trim=5 5 5 5,clip]{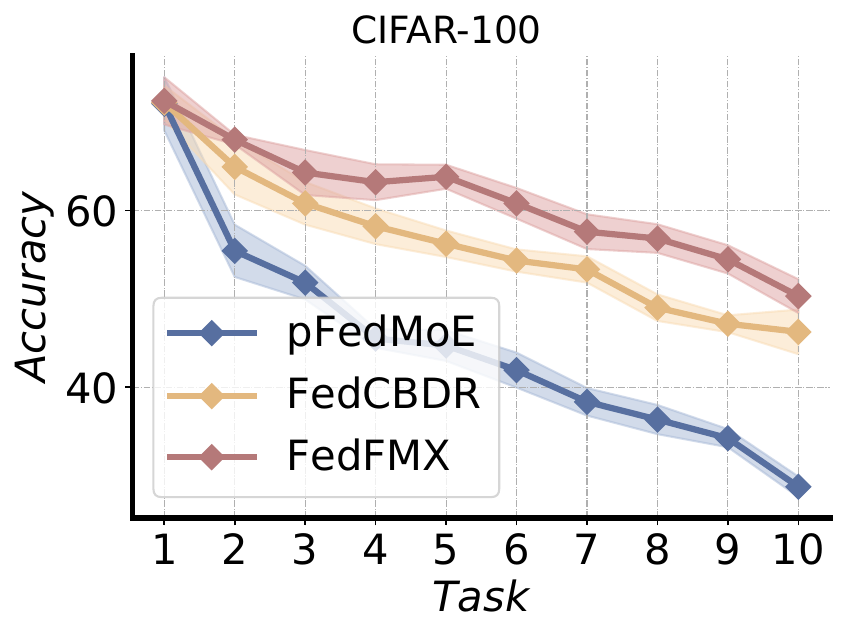}}
            \subfloat{
                \includegraphics[width=0.24\textwidth, trim=5 5 5 5,clip]{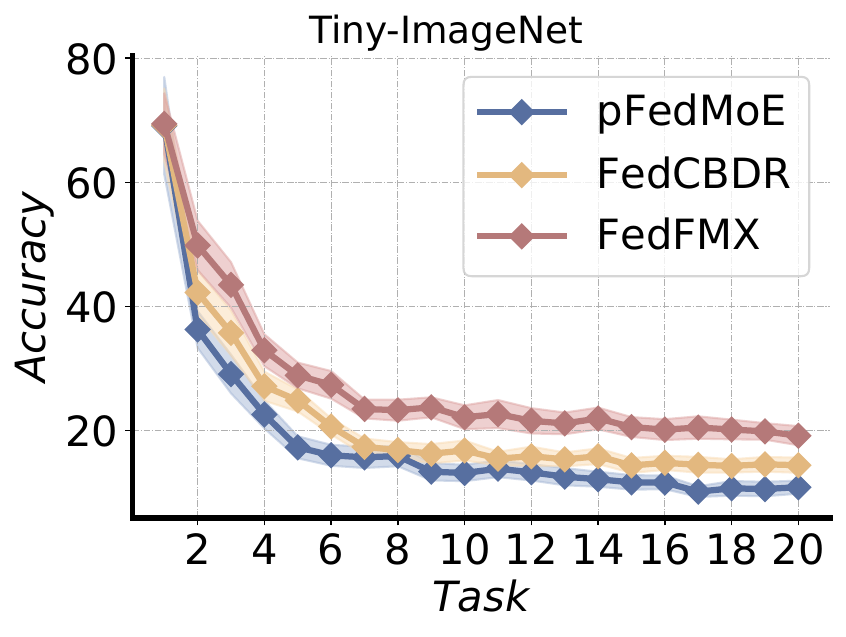}}
            \subfloat{
                \includegraphics[width=0.24\textwidth, trim=5 5 5 5,clip]{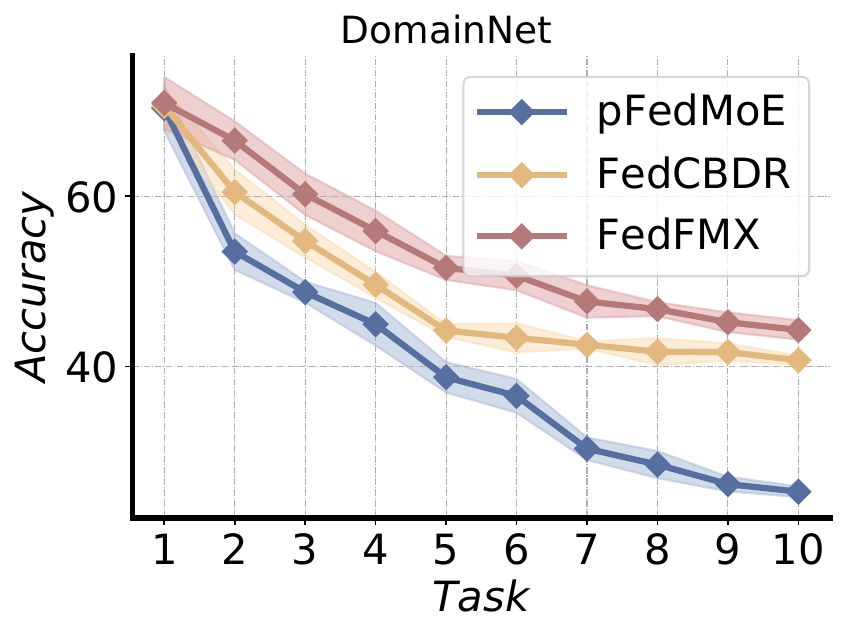}}
            \caption{Performance comparison under the fine-grained setting.} 
            \label{task_analysis_fine}
            \vspace{5pt}
        \end{minipage}
        \begin{minipage}{345pt}
            \subfloat{
                \includegraphics[width=0.24\textwidth, trim= 5 5 5 5,clip]{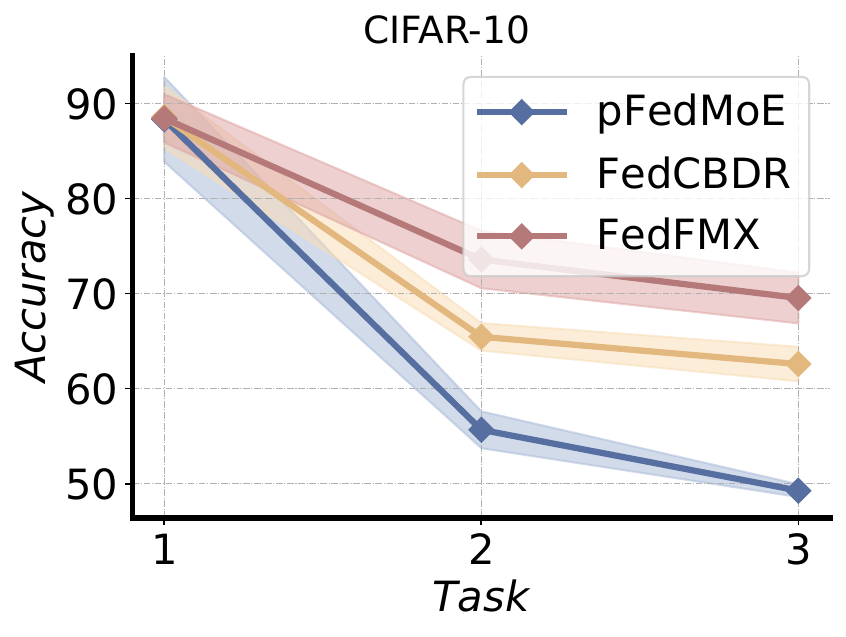}}
            \subfloat{
                \includegraphics[width=0.24\textwidth, trim=5 5 5 5,clip]{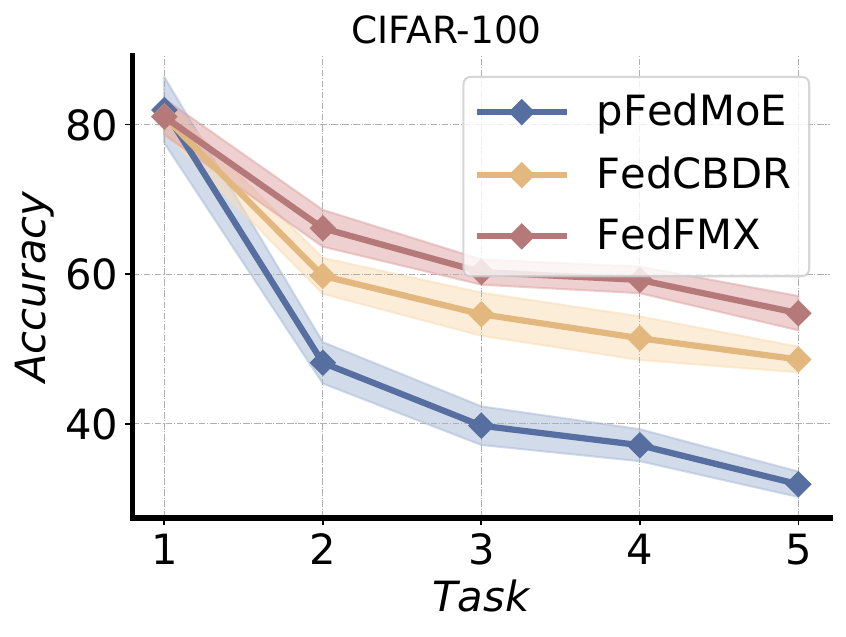}}
            \subfloat{
                \includegraphics[width=0.24\textwidth, trim=5 5 5 5,clip]{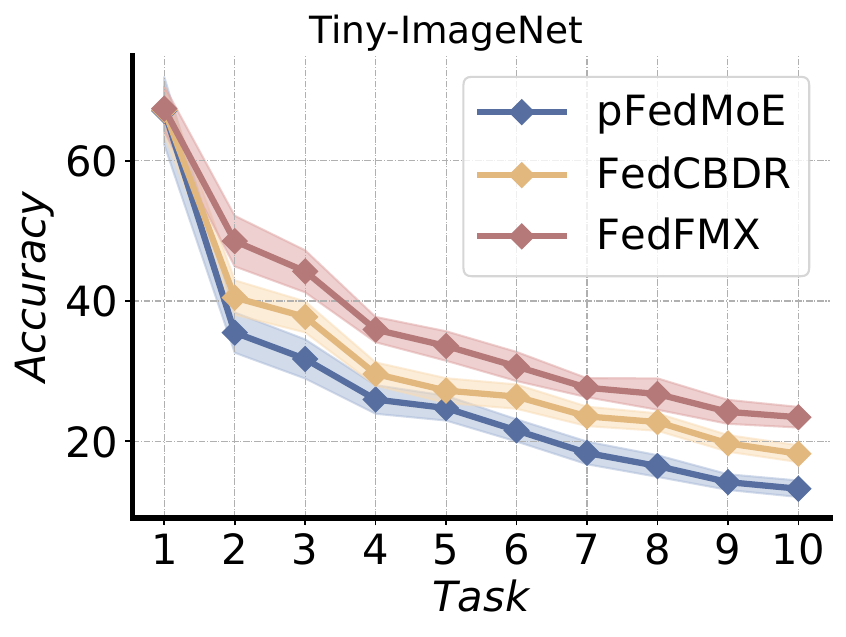}}
            \subfloat{
                \includegraphics[width=0.24\textwidth, trim=5 5 5 5,clip]{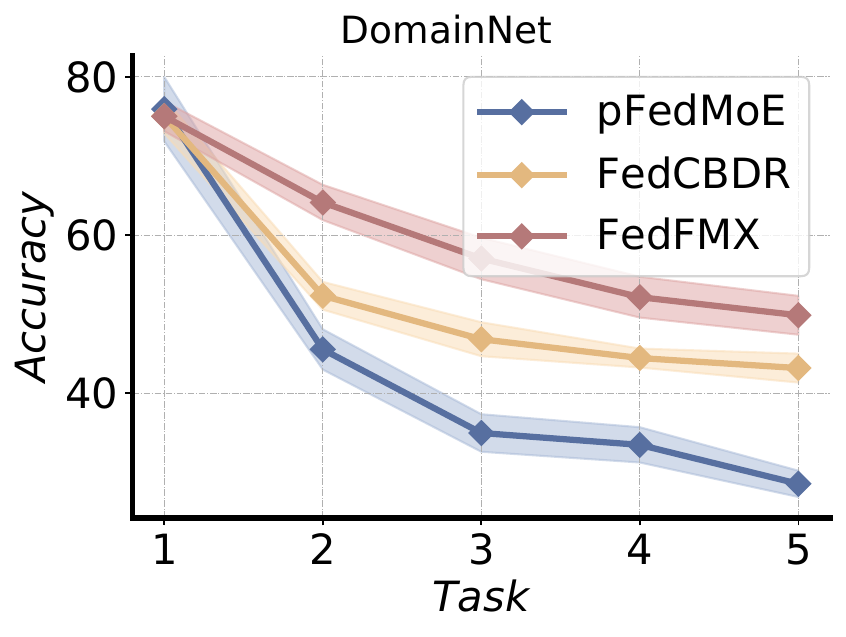}}
            \caption{Performance comparison under the coarse-grained setting.}
            \label{task_analysis_coarse}
        \end{minipage}
    \end{minipage}
\end{figure*}

\begin{table*}[t]
\centering
\renewcommand\arraystretch{1.1}
\caption{Performance comparisons of expert selection strategies during training.}
\vspace{-5pt}
\label{expert_quantity}
\resizebox{1.0\textwidth}{!}{
\begin{tabular}
{c|>{\columncolor{cifarcolor!60}}c>{\columncolor{cifarcolor!60}}c>{\columncolor{cifarcolor!60}}c|>{\columncolor{tinycolor!60}}c>{\columncolor{tinycolor!60}}c>{\columncolor{tinycolor!60}}c|>{\columncolor{domaincolor!60}}c>{\columncolor{domaincolor!60}}c>{\columncolor{domaincolor!60}}c|>{\columncolor{cifarcolor!60}}c>{\columncolor{cifarcolor!60}}c>{\columncolor{cifarcolor!60}}c|>{\columncolor{tinycolor!60}}c>{\columncolor{tinycolor!60}}c>{\columncolor{tinycolor!60}}c|>{\columncolor{domaincolor!60}}c>{\columncolor{domaincolor!60}}c>{\columncolor{domaincolor!60}}c} 
\toprule[1.2pt]

\multirow{3}{*}{\multirowcell{2}{\textbf{Non-IID} \\ \textbf{Settings}}} & \multicolumn{9}{c|}{\centering\textbf{Fine-Grained Increment}} & \multicolumn{9}{c}{\centering\textbf{Coarse-Grained Increment}}    \\ \cmidrule[0.5pt](l{1pt}r{0pt}){2-19}

& \multicolumn{3}{>{\columncolor{cifarcolor!60}}c|}{\centering\textbf{CIFAR-100}} & \multicolumn{3}{>{\columncolor{tinycolor!60}}c|}{\centering\textbf{Tiny-ImageNet}} & \multicolumn{3}{>{\columncolor{domaincolor!60}}c|}{\centering\textbf{DomainNet}} & \multicolumn{3}{>{\columncolor{cifarcolor!60}}c|}{\centering\textbf{CIFAR-100}} & \multicolumn{3}{>{\columncolor{tinycolor!60}}c|}{\centering\textbf{Tiny-ImageNet}} & \multicolumn{3}{>{\columncolor{domaincolor!60}}c}{\centering\textbf{DomainNet}}    \\ \cmidrule[0.5pt](l{1pt}r{0pt}){2-19}

& Top-$1$ & Top-$K$ & AES & Top-$1$ & Top-$K$ & AES & Top-$1$ & Top-$K$ & AES & Top-$1$ & Top-$K$ & AES & Top-$1$ & Top-$K$ & AES & Top-$1$ & Top-$K$ & AES \\ \cmidrule[0.8pt](l{1pt}r{0pt}){1-19}

QLI & 46.38 & 49.12 & \textbf{51.18} & 15.83 & 17.72 & \textbf{19.47} & 42.38 & 45.92 & \textbf{47.63} & 51.03 & 53.45 & \textbf{55.63} & 20.63 & 22.81 & \textbf{24.06} & 46.71 & 48.95 & \textbf{51.84}  \\  

DLI & 45.96 & 48.54 & \textbf{50.32} & 15.17 & 16.88 & \textbf{19.14} & 41.59 & 44.86 & \textbf{44.29} & 50.76 & 52.94 & \textbf{54.78} & 19.75 & 20.87 & \textbf{23.42} & 45.89 & 48.12 & \textbf{49.85}  \\  

\bottomrule[1.2pt]
\end{tabular}
}
\end{table*}

\subsection{Main Results and Analysis}
We evaluate the performance of \textsc{FedFMX} under two label imbalance scenarios, namely QLI and DLI, as summarized in Table~\ref{accuracy_QLI} and Table~\ref{accuracy_DLI}. Our method consistently achieves superior performance compared with benchmark methods, demonstrating the effectiveness of \textsc{FedFMX}. Under the QLI setting, \textsc{FedFMX} maintains stable performance as the degree of label imbalance increases, whereas conventional approaches tend to suffer from degraded generalization due to intensified forgetting and capacity conflicts, highlighting the advantage of the proposed Fisher-guided routing and adaptive expert selection mechanisms, which effectively coordinate expert specialization while mitigating gradient inconsistencies caused by skewed class distributions. The advantage of \textsc{FedFMX} becomes more evident on challenging datasets such as Tiny-ImageNet and DomainNet, where the number of classes and data complexity are substantially higher. In such scenarios, the proposed framework remains robust and continues to outperform competing approaches, indicating its strong capability to generalize to high-dimensional and heterogeneous federated environments.

\begin{figure*}[t]
\setlength{\abovecaptionskip}{2pt} 
    \centering
        \begin{minipage}{345pt}
            \subfloat{
                \includegraphics[width=0.32\textwidth, trim= 0 0 0 0,clip]{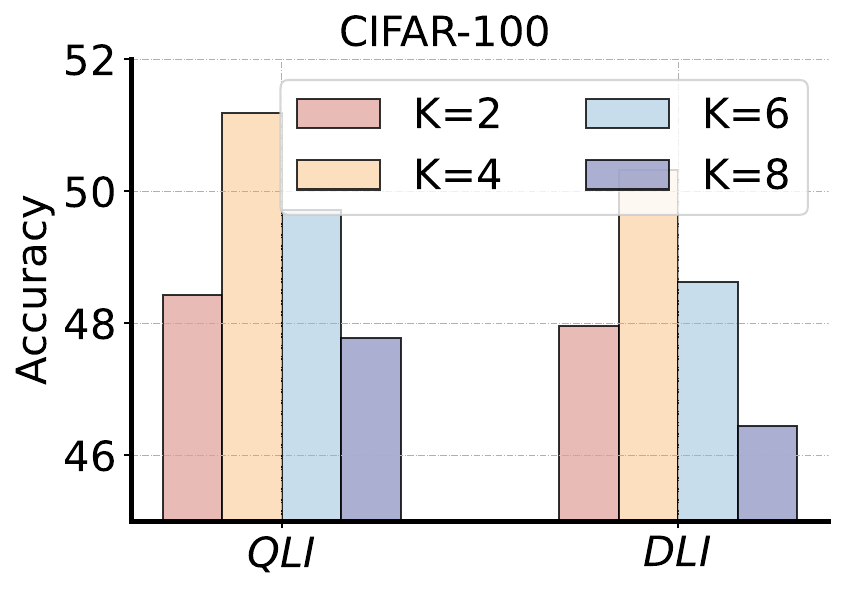}}
            \subfloat{
                \includegraphics[width=0.32\textwidth, trim=0 0 0 0,clip]{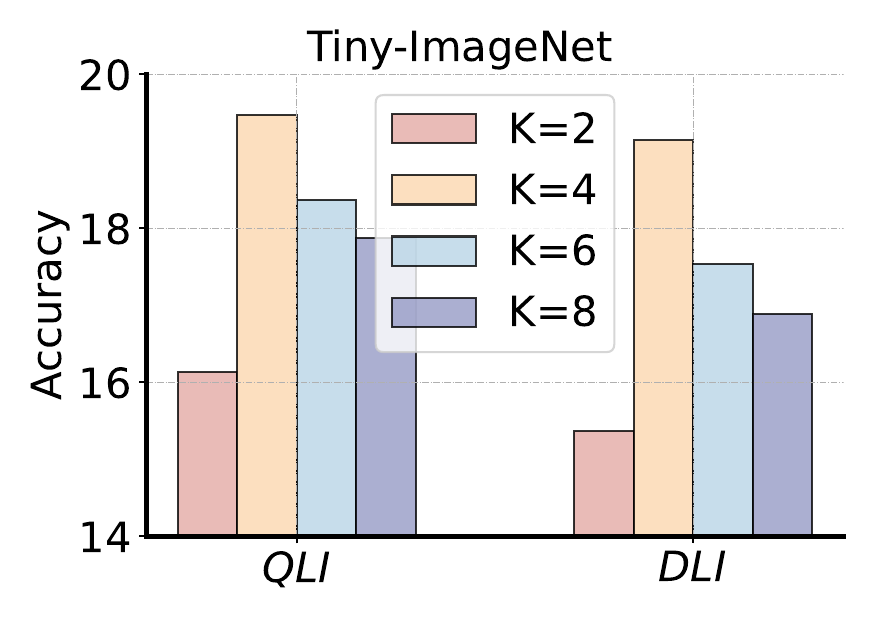}}
            \subfloat{
                \includegraphics[width=0.32\textwidth, trim=0 0 0 0,clip]{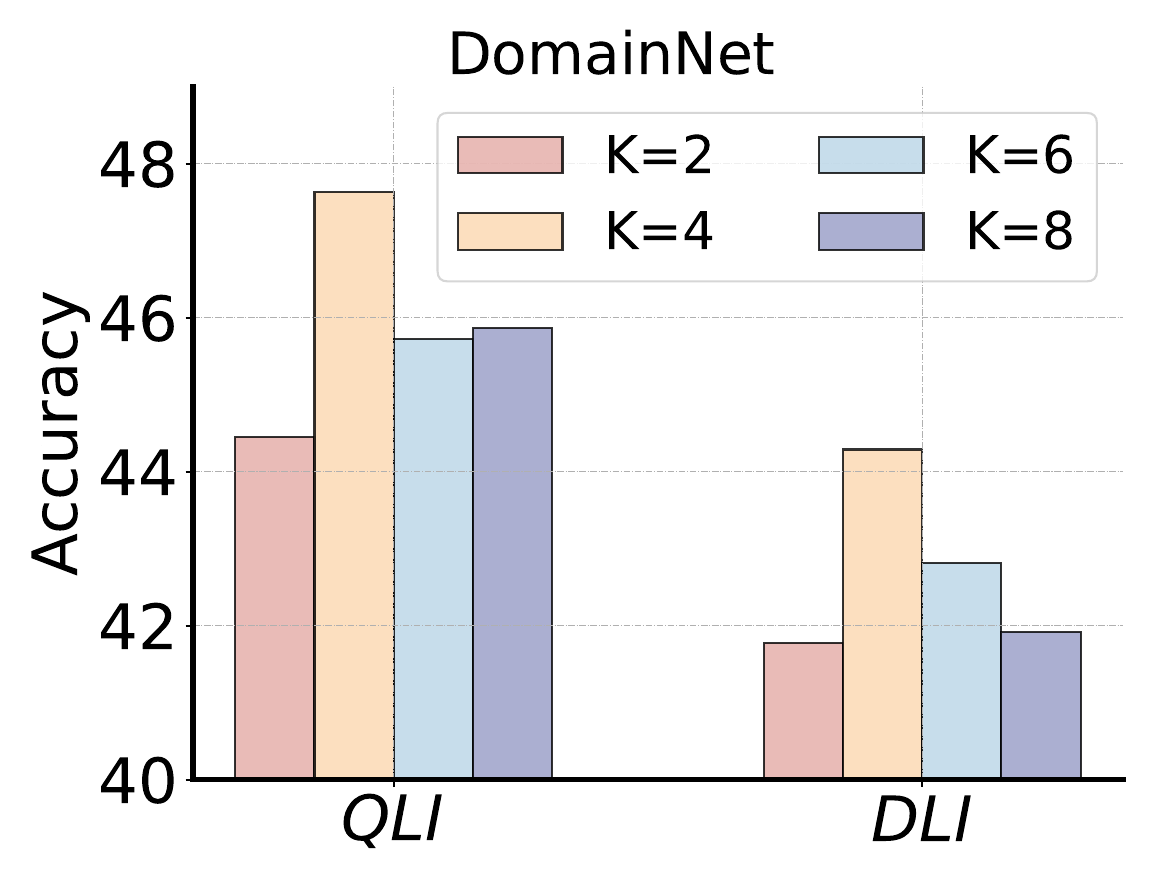}}
            \caption{Performance on the quantity of total experts under the fine-grained setting.}
            \label{expert_size_fine}
            \vspace{6pt}
        \end{minipage}
        \begin{minipage}{345pt}
            \subfloat{
                \includegraphics[width=0.32\textwidth, trim= 0 0 0 0,clip]{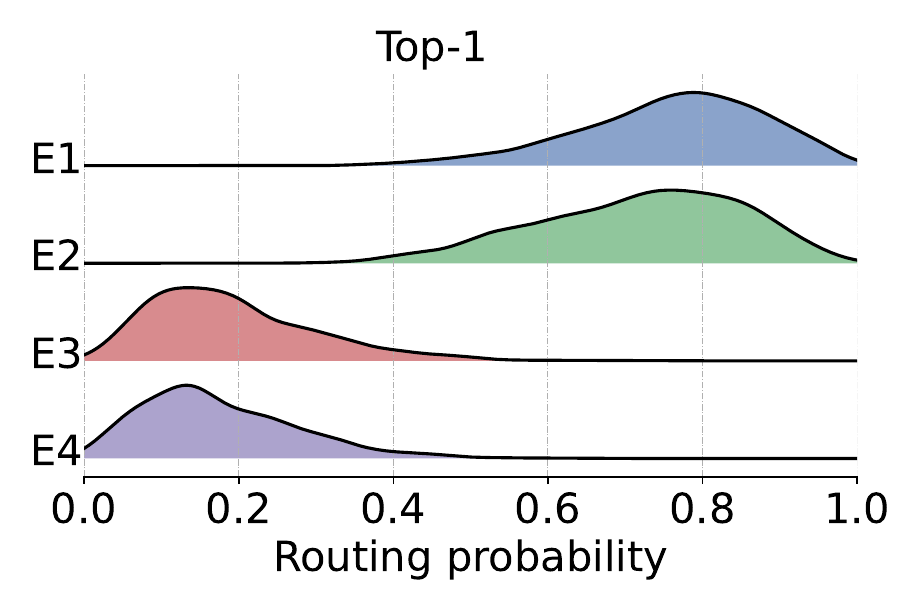}}
            \subfloat{
                \includegraphics[width=0.32\textwidth, trim=0 0 0 0,clip]{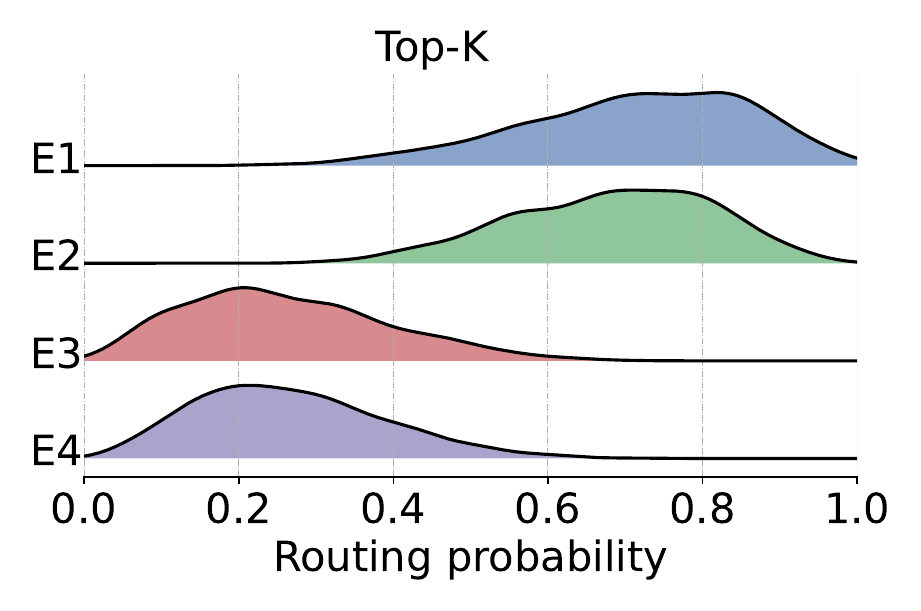}}
            \subfloat{
                \includegraphics[width=0.32\textwidth, trim=0 0 0 0,clip]{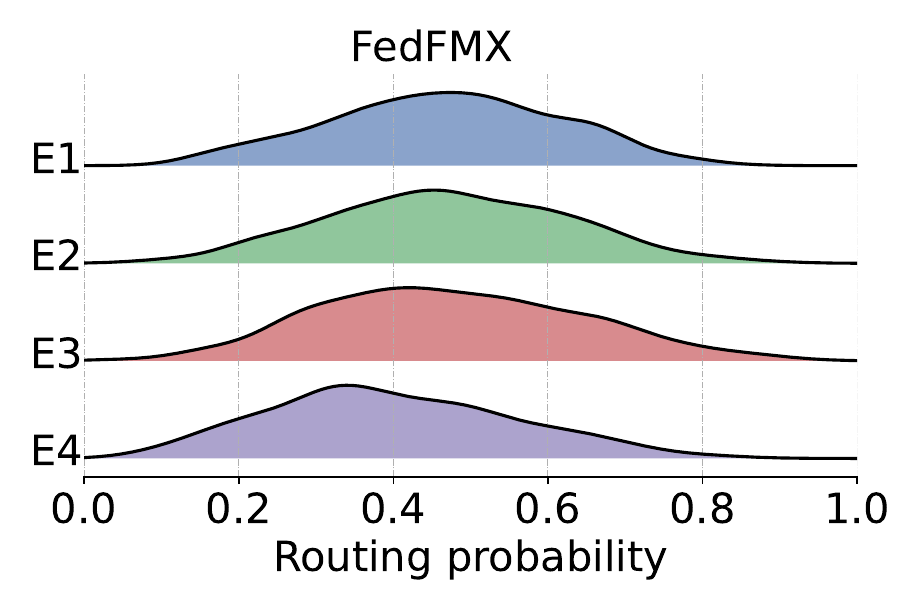}}
            \caption{Comparison of expert routing probability distributions on DomainNet.}
            \label{routing_probability}
        \end{minipage}
\end{figure*}

\subsection{Performance under Incremental Tasks} 
Figures~\ref{task_analysis_fine}-\ref{task_analysis_coarse} illustrate the performance evolution across incremental tasks under the fine- and coarse-grained settings, where $\alpha_{D}=0.5$ by default. As new tasks are introduced, the performance of all methods gradually declines due to catastrophic forgetting. Nevertheless, \textsc{FedFMX} consistently maintains higher accuracy throughout the incremental process. The performance gap between \textsc{FedFMX} and the baseline methods becomes increasingly evident as the number of tasks grows. While competing approaches exhibit noticeable degradation in later stages, \textsc{FedFMX} shows a significantly slower decline, indicating that the proposed Fisher-guided routing and adaptive expert selection strategies effectively preserve previously acquired knowledge while still enabling the model to adapt to newly introduced classes, demonstrating stronger resilience to drastic task transitions and improved stability over long incremental sequences.

\begin{figure*}[t]
\centering
\subfloat{
    \includegraphics[width=0.24\textwidth, trim= 5 5 5 5,clip]{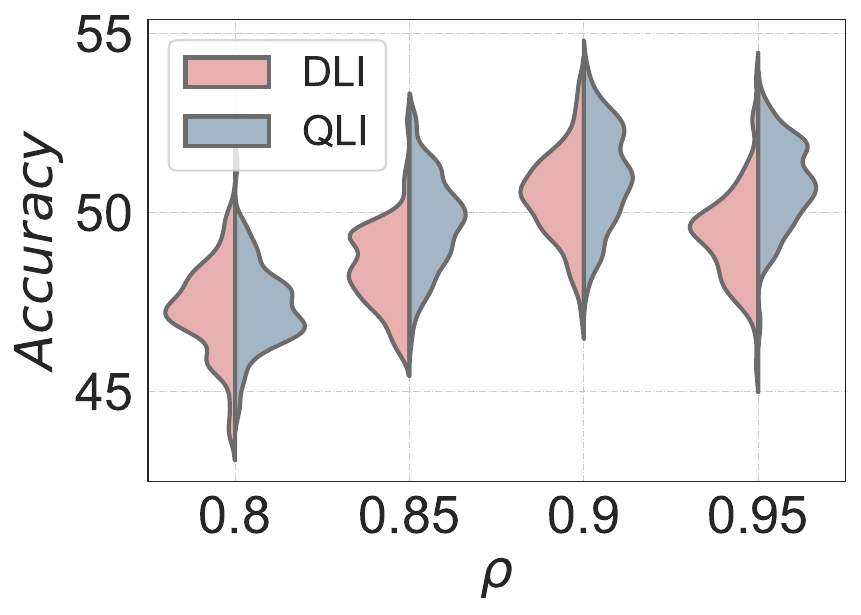}}
\subfloat{
    \includegraphics[width=0.24\textwidth, trim=5 5 5 5,clip]{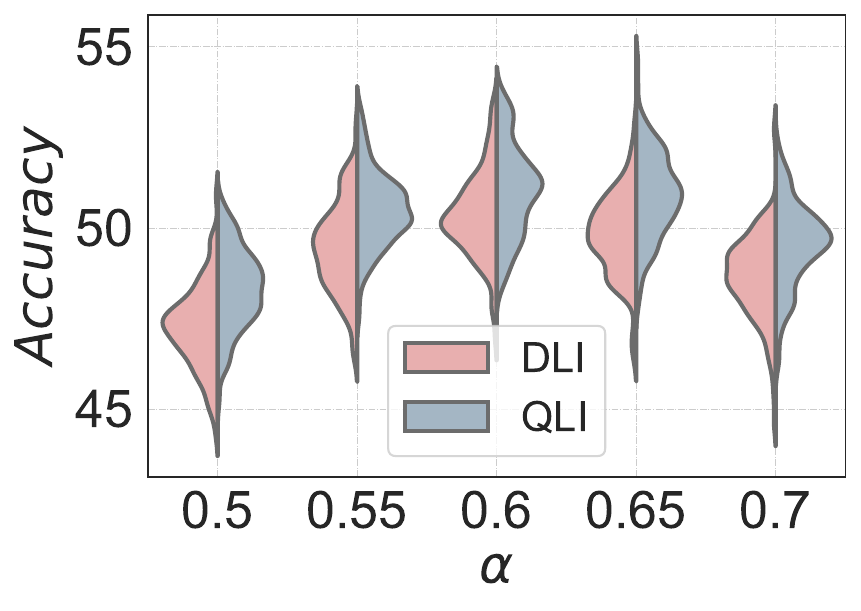}}
\subfloat{
    \includegraphics[width=0.24\textwidth, trim=5 5 5 5,clip]{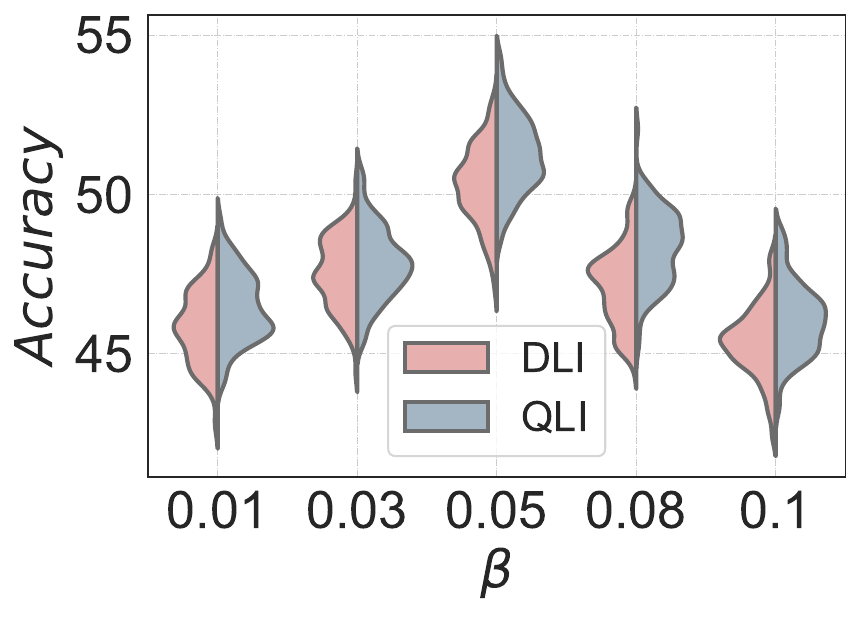}}
\subfloat{
    \includegraphics[width=0.24\textwidth, trim=5 5 5 5,clip]{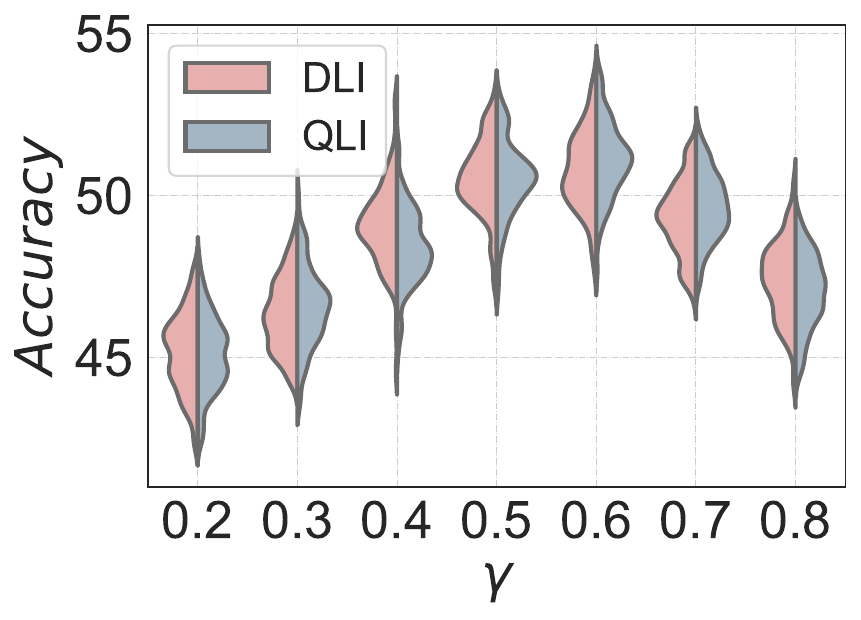}}
\vspace{-5pt}
\caption{Hyperparameters sensitivity on CIFAR-100 under the QLI/DLI setting.}
\label{hyper}
\end{figure*}

\begin{table*}[t]
\centering
\renewcommand\arraystretch{1.0}
\caption{Efficacy of each module on various datasets.}
\label{ablation}
\vspace{-18pt}
\begin{subtable}{\linewidth}
\centering
\caption{Under the QLI setting}
\vspace{-8pt}
\resizebox{1.0\textwidth}{!}{
\begin{tabular}
{ccc|>{\columncolor{cifarcolor!60}}c|>{\columncolor{tinycolor!60}}c|>{\columncolor{domaincolor!60}}c|>{\columncolor{cifarcolor!60}}c|>{\columncolor{tinycolor!60}}c|>{\columncolor{domaincolor!60}}c} 
\toprule[1.2pt]

\multicolumn{3}{c|}{\centering\textbf{Module}} & \multicolumn{3}{c|}{\centering\textbf{Fine-Grained Increment}} & \multicolumn{3}{c}{\centering\textbf{Coarse-Grained Increment}}    \\ \cmidrule[0.5pt](l{1pt}r{0pt}){1-9}

\textbf{FRES} & \textbf{AES} & \textbf{RAR} & \textbf{CIFAR-100} & \textbf{Tiny-ImageNet} & \textbf{DomainNet} & \textbf{CIFAR-100} & \textbf{Tiny-ImageNet} & \textbf{DomainNet}    \\ \cmidrule[0.5pt](l{1pt}r{0pt}){1-9}

&  &  & 33.24 & 10.86 & 29.41 & 38.57 & 13.39 & 34.18  \\ 
& \ding{52} & \ding{52}  & 48.33 & 17.96 & 45.58 & 53.41 & 22.51 & 49.92 \\  

\ding{52} &  & \ding{52} & 42.71 & 15.84 & 40.83 & 48.26 & 19.87 & 45.62 \\  

\ding{52} & \ding{52} &  & 46.23 & 16.21 & 44.12 & 51.26 & 21.63 & 48.67  \\ \midrule[0.8pt]

\ding{52} & \ding{52} & \ding{52} & \textbf{51.18} & \textbf{19.74} & \textbf{47.63} & \textbf{55.63} & \textbf{24.06} & \textbf{51.84}  \\ 

\bottomrule[1.2pt]
\end{tabular}}
\label{ablation_QLI}
\end{subtable}

\vspace{5pt}  

\begin{subtable}{\linewidth}
\centering
\caption{Under the DIL setting}
\vspace{-8pt}
\resizebox{1.0\textwidth}{!}{\begin{tabular}
{ccc|>{\columncolor{cifarcolor!60}}c|>{\columncolor{tinycolor!60}}c|>{\columncolor{domaincolor!60}}c|>{\columncolor{cifarcolor!60}}c|>{\columncolor{tinycolor!60}}c|>{\columncolor{domaincolor!60}}c} 
\toprule[1.2pt]

\multicolumn{3}{c|}{\centering\textbf{Module}} & \multicolumn{3}{c|}{\centering\textbf{Fine-Grained Increment}} & \multicolumn{3}{c}{\centering\textbf{Coarse-Grained Increment}}    \\ \cmidrule[0.5pt](l{1pt}r{0pt}){1-9}

\textbf{FRES} & \textbf{AES} & \textbf{RAR} & \textbf{CIFAR-100} & \textbf{Tiny-ImageNet} & \textbf{DomainNet} & \textbf{CIFAR-100} & \textbf{Tiny-ImageNet} & \textbf{DomainNet}    \\ \cmidrule[0.5pt](l{1pt}r{0pt}){1-9}

&  &  & 32.11 & 10.42 & 26.83 & 37.69 & 12.94 & 31.88  \\ 
& \ding{52} & \ding{52}  & 47.98 & 17.41 & 42.71 & 52.89 & 21.88 & 47.83 \\  

\ding{52} &  & \ding{52} & 41.86 & 15.32 & 38.34 & 47.63 & 19.46 & 44.02 \\  

\ding{52} & \ding{52} &  & 45.71 & 16.76 & 41.63 & 50.91 & 21.17 & 47.12   \\ \midrule[0.8pt]

\ding{52} & \ding{52} & \ding{52} & \textbf{50.32} & \textbf{19.14} & \textbf{44.29} & \textbf{54.78} & \textbf{23.42} & \textbf{49.85}  \\ 

\bottomrule[1.2pt]
\end{tabular}
}
\label{ablation_DLI}
\end{subtable}
\end{table*}

\subsection{Expert Quantity Study}
To validate the effectiveness of our AES module, we compare with Top-$1$ and Top-$K$ routing strategies under both QLI/DLI and fine-/coarse-grained incremental settings. As depicted in Table~\ref{expert_quantity}, AES consistently achieves the best performance across all datasets and settings. Compared with Top-$1$ routing, which activates only a single expert, AES provides improved performance by enabling multiple experts to collaboratively capture diverse task-specific representations. Meanwhile, although Top-$K$ routing increases the number of activated experts, its fixed selection strategy cannot adapt to the varying importance of experts across different tasks and data distributions. In contrast, AES dynamically determines the expert subset based on the learned importance signals, allowing the model to better balance expert specialization and knowledge sharing.

To investigate the influence of expert quantity on model performance, we vary the expert pool size $K \in \{2, 4, 6, 8\}$ under both QLI and DLI scenarios with $\alpha_{D} = 0.5$ and the fine-grained incremental task setting, as shown in Figure~\ref{expert_size_fine}. The results indicate that increasing the quantity of experts initially improves the performance, suggesting that a moderate expansion of expert capacity enables the capture of diverse task-specific knowledge. However, with an excessively large expert pool, performance decreases, implying that although a larger number of experts increases representational capacity, it may also introduce stronger routing competition and knowledge fragmentation, thereby weakening the effectiveness of expert specialization. Consequently, the moderate quantity of experts achieves a better balance between model capacity and expert coordination. 

To further study the expert utilization under different routing strategies, in Figure~\ref{routing_probability}, we analyze the routing probability distributions of individual experts on DomainNet under Top-$1$/Top-$K$ routing, and \textsc{FedFMX}. Top-$1$ routing leads to highly skewed expert utilization, where only a small subset of experts receives high routing probabilities while others are rarely activated, limiting expert diversity and leading to capacity imbalance. Top-$K$ routing partially alleviates this issue, but the routing probabilities remain biased, resulting in uneven expert participation. In contrast, \textsc{FedFMX} produces a noticeably more balanced routing distribution, indicating that different experts are utilized more consistently, which suggests that our method effectively promotes expert diversity while mitigating expert collapse, and better exploits the capacity of the expert pool.

\begin{table*}[t]
\centering
\renewcommand\arraystretch{1.0}
\caption{Computational cost analysis under different backbone architectures.}
\label{communication}
\vspace{-18pt}
\begin{subtable}{\linewidth}
\centering
\caption{ResNet-18}
\vspace{-8pt}
\resizebox{1.0\textwidth}{!}{
\begin{tabular}
{c|>{\columncolor{cifarcolor!60}}c>{\columncolor{cifarcolor!60}}c>{\columncolor{cifarcolor!60}}c>{\columncolor{cifarcolor!60}}c|>{\columncolor{tinycolor!60}}c>{\columncolor{tinycolor!60}}c>{\columncolor{tinycolor!60}}c>{\columncolor{tinycolor!60}}c|>{\columncolor{domaincolor!60}}c>{\columncolor{domaincolor!60}}c>{\columncolor{domaincolor!60}}c>{\columncolor{domaincolor!60}}c} 
\toprule[1.2pt]

\multirow{1.5}{*}{\multirowcell{2}{\textbf{Method}}} & \multicolumn{4}{>{\columncolor{cifarcolor!60}}c|}{\centering\textbf{CIFAR-100}} & \multicolumn{4}{>{\columncolor{tinycolor!60}}c|}{\centering\textbf{Tiny-ImageNet}} & \multicolumn{4}{>{\columncolor{domaincolor!60}}c}{\centering\textbf{DomainNet}}  \\ \cmidrule[0.5pt](l{1pt}r{0pt}){2-13}

& Params & FLOPs & Time(s) & Acc & Params & FLOPs & Time(s) & Acc & Params & FLOPs & Time(s) & Acc \\ \cmidrule[0.8pt](l{1pt}r{0pt}){1-13}

\textsc{pFedMoE} & 11.7M & 89 & 75 & 30.76 & 11.9M & 487 & 375 & 11.02 & 12.0M & 1573 & 826 & 27.92 \\  

\textsc{FedFMX} & 14.6M & 98 & 82 & 51.18 & 14.7M & 588 & 412 & 19.47 & 15.0M & 1785 & 934 &  47.63 \\  

\bottomrule[1.2pt]
\end{tabular}}
\label{overhead_QLI}
\end{subtable}

\vspace{5pt}  

\begin{subtable}{\linewidth}
\centering
\caption{ViT-B/16}
\vspace{-8pt}
\resizebox{1.0\textwidth}{!}{
\begin{tabular}
{c|>{\columncolor{cifarcolor!60}}c>{\columncolor{cifarcolor!60}}c>{\columncolor{cifarcolor!60}}c>{\columncolor{cifarcolor!60}}c|>{\columncolor{tinycolor!60}}c>{\columncolor{tinycolor!60}}c>{\columncolor{tinycolor!60}}c>{\columncolor{tinycolor!60}}c|>{\columncolor{domaincolor!60}}c>{\columncolor{domaincolor!60}}c>{\columncolor{domaincolor!60}}c>{\columncolor{domaincolor!60}}c} 
\toprule[1.2pt]

\multirow{1.5}{*}{\multirowcell{2}{\textbf{Method}}} & \multicolumn{4}{>{\columncolor{cifarcolor!60}}c|}{\centering\textbf{CIFAR-100}} & \multicolumn{4}{>{\columncolor{tinycolor!60}}c|}{\centering\textbf{Tiny-ImageNet}} & \multicolumn{4}{>{\columncolor{domaincolor!60}}c}{\centering\textbf{DomainNet}}  \\ \cmidrule[0.5pt](l{1pt}r{0pt}){2-13}

& Params & FLOPs & Time(s) & Acc & Params & FLOPs & Time(s) & Acc & Params & FLOPs & Time(s) & Acc \\ \cmidrule[0.8pt](l{1pt}r{0pt}){1-13}

\textsc{pFedMoE} & 89.5M & 902 & 645  & 45.37 & 89.7M & 928 & 647 & 22.51 & 89.8M & 928 & 692 & 41.25  \\  

\textsc{FedFMX} & 93.0M & 974  & 696  & 68.44 & 93.2M & 984 & 716 & 30.92 & 93.3M & 1015 & 759 & 63.94    \\  

\bottomrule[1.2pt]
\end{tabular}
}
\label{overhead_DLI}
\end{subtable}
\end{table*}

\subsection{Hyerparameter Sensitivity Analysis}
We evaluate the robustness of our method regarding hyperparameter choices and conduct a sensitivity analysis on CIFAR-100 under the DLI/QLI setting by varying four key parameters. As illustrated in Fig.~6, \textsc{FedFMX} maintains relatively stable performance across a wide range of parameter values. Specifically, for $\rho$, the performance first improves and then slightly decreases as $\rho$ increases, indicating that a moderate sparsity level provides a better balance between model capacity and structural regularization. Similar trends can be observed for $\alpha$ and $\beta$. The results demonstrate that \textsc{FedFMX} achieves robust performance without requiring delicate hyperparameter tuning, highlighting the stability and practical applicability of the proposed framework.

\subsection{Ablation Studies}
In Table~\ref{ablation}, we evaluate the contribution of each component in \textsc{FedFMX} and conduct an ablation study by progressively enabling the three key modules. Each module consistently improves the overall performance across different datasets and incremental settings. Removing any component leads to noticeable performance degradation, indicating complementary benefits. In particular, FRES improves expert routing quality by guiding the model to identify stable and informative experts, while AES further enhances the routing process by adaptively selecting expert subsets based on their estimated importance. Meanwhile, RAR introduces additional regularization that encourages more balanced expert utilization and stabilizes training during incremental updates. When all components are jointly enabled, the model achieves the best performance across both QLI and DLI settings as well as across fine-grained and coarse-grained increments. These results demonstrate that the three modules cooperate effectively to improve expert routing, mitigate knowledge interference, and enhance robustness.

\subsection{Communication Studies}
To evaluate the computational efficiency, we compare \textsc{FedFMX} with \textsc{pFedMoE} under two backbone architectures under the fine-grained incremental setting. As shown in Table~\ref{communication}, \textsc{FedFMX} introduces only a modest increase in computational cost compared with \textsc{pFedMoE}. Despite this overhead, \textsc{FedFMX} consistently achieves significantly better model performance, which is particularly notable on challenging datasets, such as Tiny-ImageNet and DomainNet. These results indicate that the proposed routing and expert selection mechanisms effectively improve model capacity and knowledge retention while maintaining a reasonable computational overhead, demonstrating a favorable trade-off between efficiency and performance.

\section{Conclusion}
In this paper, we presented \textsc{FedFMX}, a Mixture-of-Experts-based framework tailored for FCIL. To address the fundamental challenges of catastrophic forgetting, data heterogeneity, and synchronized class misalignment, \textsc{FedFMX} introduces a Fisher-Routed Expert Scoring module that quantifies each expert’s stability–plasticity trade-off, an Adaptive Expert Selection module based on marginal contributions to dynamically activate expert subsets, and a routing-aware regularization scheme that balances expert utilization and supports efficient inference. By unifying these components, \textsc{FedFMX} enables stable incremental learning, expert specialization, and flexible deployment without relying on ground-truth labels during inference. Rigorous theoretical analyses prove the $\mathcal{O}(T^{-1})$ convergence rate. Experiments across multiple benchmarks demonstrate that our method consistently outperforms state-of-the-art approaches, validating the effectiveness for realistic and scalable FCIL scenarios.


\section*{Acknowledgements}
This work was supported by the UGC General Research Fund no. 17209822 and the Innovation and Technology Commission Fund no. ITS/383/23FP from Hong Kong.

%
%
\bibliographystyle{splncs04}
\bibliography{reference}

\end{document}